\documentclass[10pt,journal]{IEEEtran}

\ifCLASSOPTIONcompsoc

  \usepackage[nocompress]{cite}
\else
  \usepackage{cite}
\fi

\usepackage{multirow}
\usepackage{makecell}
\usepackage{graphicx}
\usepackage{times}
\usepackage{epsfig}
\usepackage{amsmath}
\usepackage{amssymb}
\usepackage{multicol}
\usepackage{balance}
\usepackage{multirow}
\usepackage{float}
\usepackage{caption}
\usepackage{ctable}
\usepackage{array}
\usepackage{makecell}
\usepackage{subfig}
\usepackage{tabularx}
\usepackage{bbding}
\usepackage{pifont}
\usepackage{url}
\usepackage{times}
\usepackage{epsfig}
\usepackage{amsmath}
\usepackage{amssymb}
\usepackage{colortbl}
\usepackage{color}
\usepackage{threeparttable}
\usepackage{booktabs}
\usepackage{adjustbox}
\usepackage{subfig}
\usepackage{xfrac}
\usepackage{tabto,enumitem}
\usepackage[linesnumbered,lined,boxed,commentsnumbered,ruled]{algorithm2e}
\usepackage[breaklinks=true,colorlinks,citecolor=blue,urlcolor=blue,linkcolor=blue,bookmarks=false]{hyperref} % pagebackref

\usepackage{xspace}

\usepackage{xspace}
\makeatletter
\DeclareRobustCommand\onedot{\futurelet\@let@token\@onedot}
\def\@onedot{\ifx\@let@token.\else.\null\fi\xspace}

\begin{document}

\title{Change Detection Methods for Remote Sensing in the Last Decade: A Comprehensive Review}
\newcommand{\xt}[1]{{\color{red}(xiangtai: {#1})}} % xiangtai's comments
\newcommand{\gl}[1]{{\color{blue}(guangliang: {#1})}} % guangliang's comments
\author{Guangliang Cheng$^*$, Yunmeng Huang$^*$, Xiangtai Li, Shuchang Lyu, Zhaoyang Xu, Qi Zhao, Shiming Xiang,
        \thanks {Guangliang Cheng is with the Department of Computer Science, at the University of Liverpool. \{Guangliang.Cheng\}@liverpool.ac.uk.}% <-this % stops a space
        \thanks{Yunmeng Huang, Shuchang Lyu, and Qi Zhao are with Beihang University. \{yunmenghuang, lyushuchang, zhaoqi\}@buaa.edu.cn}
        \thanks{Xiangtai Li is with Peking Univeristy. \{lxtpku\}@pku.edu.cn.}
        \thanks{Zhaoyang Xu is with the Cambridge University.}
        % \thanks{Lefei Zhang is with Wuhan University.}
        \thanks{Shiming Xiang is with the State Key Laboratory of Multimodal Artificial Intelligence Systems, Institute of Automation, Chinese Academy of Sciences. \{smxiang\}@nlpr.ia.ac.cn}
        \thanks{Corresponding author: Guangliang Cheng, Qi Zhao.} 
        \thanks{The first two authors contribute equally.}
}

% The paper headers
\markboth{Journal of \LaTeX\ Class Files,~Vol.~XX, No.~X, XX~XXXX}%
{Cheng \MakeLowercase{\textit{et al.}}: Bare Demo of IEEEtran.cls for IEEE Journals}

\maketitle

\IEEEpeerreviewmaketitle

%\IEEEtitleabstractindextext{
\begin{abstract} % 1/4 page
\par Change detection is an essential and widely utilized task in remote sensing that aims to detect and analyze changes occurring in the same geographical area over time, which has broad applications in urban development, agricultural surveys, and land cover monitoring. Detecting changes in remote sensing images is a complex challenge due to various factors, including variations in image quality, noise, registration errors, illumination changes, complex landscapes, and spatial heterogeneity. In recent years, deep learning has emerged as a powerful tool for feature extraction and addressing these challenges. Its versatility has resulted in its widespread adoption for numerous image-processing tasks. This paper presents a comprehensive survey of significant advancements in change detection for remote sensing images over the past decade. We first introduce some preliminary knowledge for the change detection task, such as problem definition, datasets, evaluation metrics, and transformer basics, as well as provide a detailed taxonomy of existing algorithms from three different perspectives:  \textit{algorithm granularity}, \textit{supervision modes}, and \textit{learning frameworks} in the methodology section. This survey enables readers to gain systematic knowledge of change detection tasks from various angles. We then summarize the state-of-the-art performance on several dominant change detection datasets, providing insights into the strengths and limitations of existing algorithms. Based on our survey, some future research directions for change detection in remote sensing are well identified. This survey paper will shed some light on the community and inspire further research efforts in the change detection task.

%\par Change detection is a critical task in remote sensing. It aims to identify and analyze changes occurring in the same area over time. Detecting image changes poses a challenge due to various factors, such as variations in quality, noise, registration errors, illumination changes, complex landscapes, and spatial heterogeneity. In recent years, deep learning has emerged as a highly effective tool for feature extraction and addressing the aforementioned challenges. Its versatility has resulted in its widespread adoption for numerous image-processing tasks. This work presents a comprehensive survey of representative advancements in change detection for remote sensing images over the past decade. Specifically, we review the fundamental knowledge necessary for the change detection task, followed by a detailed taxonomy of existing algorithms from three different perspectives: \textit{algorithm granularity}, \textit{supervision modes}, and \textit{learning frameworks} in the methodology section. Thus, readers can gain systematic knowledge of change detection tasks from different aspects. Subsequently, we systematically summarize the state-of-the-art performance on several dominant change detection datasets. We identify future research directions in this area based on the above survey. This survey paper will shed some light on the community and inspire further research efforts in the change detection task.

\end{abstract}
%}
\section{Introduction} % 1.5 pages 
  \label{intro}
   \IEEEPARstart{C}{hange} detection in remote sensing is the process of identifying changes in a scene from a pair of images captured in the same geographical area but at different time periods. With the rapid development of earth observation technologies, acquiring large amounts of remote sensing images has become increasingly accessible. Consequently, change detection has emerged as a popular and fundamental task in the remote sensing community, with numerous real-world applications, such as urban development~\cite{DBLP:journals/tits/BuchVO11, DBLP:journals/lgrs/LiuPZZY21}, agricultural surveys~\cite{DBLP:journals/tgrs/BruzzoneF00, DBLP:journals/staeors/LiuCDL22}, land cover monitoring~\cite{DBLP:journals/tgrs/ShiZZLLZ22, DBLP:journals/tgrs/KhanHPB17, DBLP:journals/lgrs/GaoWGDW19}, disaster assessment~\cite{DBLP:conf/igarss/BrunnerBL10} and military~\cite{DBLP:journals/tnn/GongZLMJ16}.
   \begin{figure}
  \centering
  \includegraphics[width=0.48\textwidth]{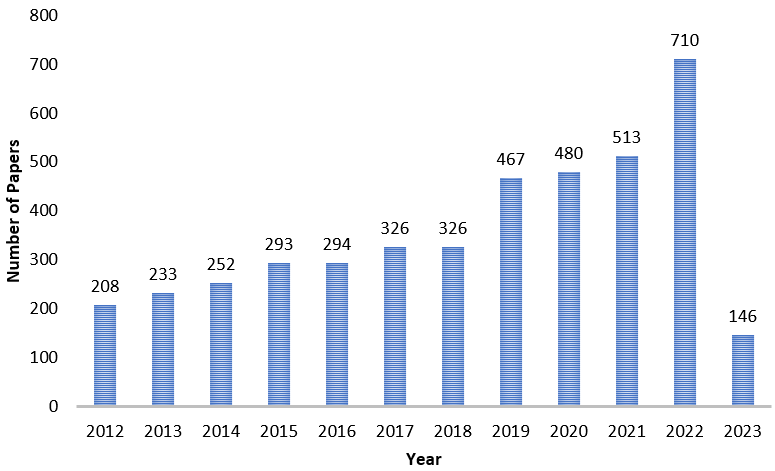}
  \caption{Statistics on the literature of change detection in the past decade.}
  \label{fig:paper_summary}
\end{figure}

  \par However, change detection poses some significant challenges as paired images are often captured under varying conditions, such as different angles, illuminations, and even during different seasons, resulting in diverse and unknown changes in a scene. These challenges encompass a broad range of issues, including 1) variations in image quality arising from differences in spatial, spectral, and temporal resolutions; 2) the presence of diverse types of noise and artifacts; 3) errors during image registration; 4) difficulties in handling illumination, shadows, and changes in viewing angle; 5) complex and dynamic landscapes; 6) scale and spatial heterogeneity of the landscape.

  \par Over the past few decades, numerous change detection methods have been proposed. Before the deep learning era, pointwise classification methods~\cite{DBLP:journals/staeors/HuoCDZP16,DBLP:journals/tgrs/BovoloBM08, DBLP:journals/ijgi/SeoKELP18, DBLP:conf/grmse/XieWHY16, doi:10.1080/01431161.2014.951740, DBLP:journals/lgrs/HaoSZL14, DBLP:journals/lgrs/LiLZSA015, DBLP:journals/lgrs/HaoZJS20, DBLP:journals/staeors/ZhouCLS16,DBLP:journals/tgrs/BruzzoneF00, DBLP:journals/tip/TouatiMD20} witnessed great progress in change detection tasks. Most traditional methods focus on detecting the changed pixels and classifying them to generate a change map. A variety of machine learning models have been applied to the change detection task, including support vector machine (SVM)~\cite{DBLP:journals/staeors/HuoCDZP16, DBLP:journals/tgrs/BovoloBM08}, random forests~\cite{DBLP:journals/ijgi/SeoKELP18}, decision trees~\cite{DBLP:conf/grmse/XieWHY16}, level set~\cite{doi:10.1080/01431161.2014.951740, DBLP:journals/lgrs/HaoSZL14}, Markov random fields (MRF)~\cite{DBLP:journals/tgrs/BruzzoneF00, DBLP:journals/tip/TouatiMD20}, and conditional random fields (CRF)~\cite{DBLP:journals/lgrs/LiLZSA015, DBLP:journals/lgrs/HaoZJS20, DBLP:journals/staeors/ZhouCLS16}. Though some specific images can achieve considerable performance with the above methods, they still suffer from low accuracy and lack of generalization ability. Moreover, their performance is highly dependent on the decision classifiers and threshold settings.

  \par In recent years, the rapid development of deep learning technologies, particularly deep convolutional neural networks (CNNs), has led to the emergence of CNN models that have shown superior performance over traditional methods for change detection tasks, as evidenced by numerous studies~\cite{DBLP:journals/lgrs/ZhanFYSWQ17, DBLP:journals/ijon/0006ZXZF21, DBLP:journals/tnn/LiuGQZ18, DBLP:journals/tnn/GongZLMJ16, DBLP:journals/tgrs/HouLWW20, DBLP:journals/lgrs/KhuranaS20, DBLP:journals/staeors/KernerWBGBA19, DBLP:journals/tgrs/ZhaoMCBE20, DBLP:journals/lgrs/LinLFG20}. This is mainly attributed to the exceptional representation and nonlinear characterization capabilities of CNNs, which make them a more effective choice for obtaining optimal performance in this field.

  \begin{figure*}
  \centering
  \includegraphics[width=1.0\textwidth]{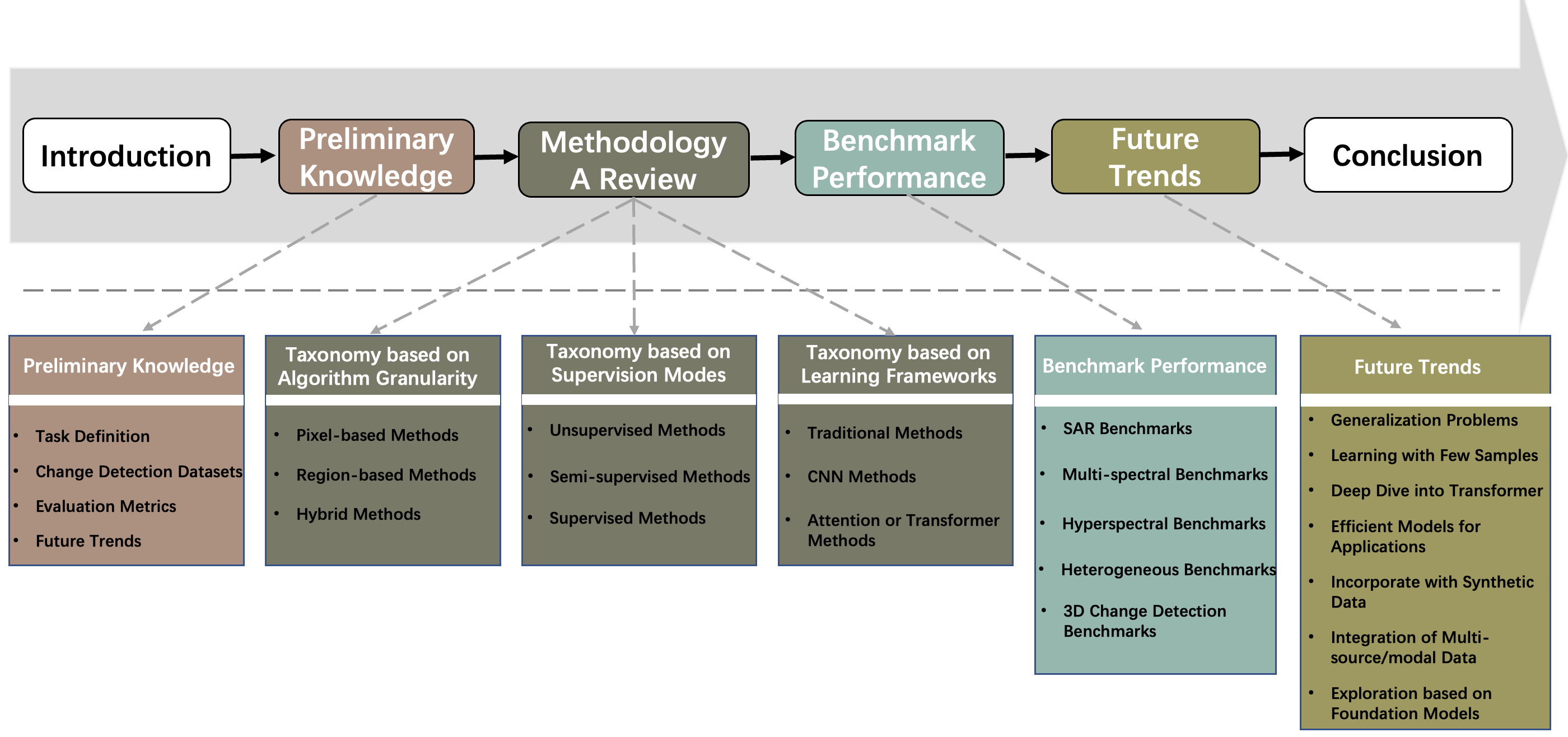}
  \caption{Illustration of the pipeline of this survey. Different colors represent specific sections. Best viewed in color.}
  \label{fig:attention_transformer}
\end{figure*}
Recently, attention models~\cite{DBLP:conf/nips/VaswaniSPUJGKP17, DBLP:journals/remotesensing/WangJLXL21, DBLP:journals/tgrs/GongJQLZLZZ22, DBLP:journals/tgrs/QuHDLX22, DBLP:journals/lgrs/MengGDDL22, DBLP:journals/tgrs/WangL22b} have been proposed to capture spatial and temporal dependencies within image pairs for change detection tasks. These models leverage an attention mechanism to focus on crucial areas, enabling them to better identify subtle changes in the scene and distinguish them from usual scene variability. On the other hand, transformer models~\cite{DBLP:conf/iclr/DosovitskiyB0WZ21, DBLP:journals/corr/abs-2207-09240, DBLP:journals/tgrs/LiZDD22, DBLP:journals/staeors/SongHL22, DBLP:journals/tgrs/ChenQS22, DBLP:journals/lgrs/SongZLSP22, DBLP:journals/staeors/LiuCDL22, DBLP:journals/lgrs/DingLZ22, DBLP:journals/corr/abs-2212-04869} employ a self-attention mechanism to learn global relationships between the image pixels, allowing them to capture long-term dependencies and spatial correlations. They have shown promising results in tasks that require modeling temporal sequences, such as video processing~\cite{DBLP:journals/csur/KhanNHZKS22} and language translation~\cite{DBLP:conf/interspeech/GangiNT19}. Both attention models and transformer models have shown significant improvements over traditional methods and Vanilla CNNs in change detection tasks, making them promising avenues for further research and development in the field of remote sensing.
%\xt{the difference with the previous survey is not clear.}
%\xt{the advantages of our survey over others must be list in more detailed or even in a table!!!}
\par With the plethora of recent change detection methods that rely on deep learning strategies, many works have emerged to survey these methods. Some of these works consider deep learning methods in general~\cite{DBLP:journals/remotesensing/YouCZ20, DBLP:journals/ecoi/AfaqM21, DBLP:journals/remotesensing/JiangPZXHLMH22, doi:10.1080/10095020.2022.2128902, DBLP:journals/remotesensing/ShafiqueCKAA22, STILLA2023228, DBLP:journals/remotesensing/LiZGZX22}, while they focus on specific aspects, such as multi-source remote sensing images and multi-objective scenarios~\cite{DBLP:journals/remotesensing/YouCZ20}, high-resolution remote sensing images~\cite{DBLP:journals/remotesensing/JiangPZXHLMH22}, and 3D point cloud data~\cite{STILLA2023228}, multi-scale attention methods~\cite{DBLP:journals/remotesensing/LiZGZX22} etc. Moreover, it is worth noting that the most recent survey method only accounts for algorithms up to 2021. Fig.~\ref{fig:paper_summary} depicts the statistical analysis of published papers related to the change detection algorithms over the past decade, as recorded in the DBLP\footnote{\url{https://dblp.org/}}. Based on the findings, we can conclude that the topic will be further explored with deep learning and an increasing number of research works will be published in the forthcoming years. However, in light of the increased focus on attention and transformer methods~\cite{DBLP:journals/corr/abs-2201-05047, DBLP:journals/csur/KhanNHZKS22, Transformer_survey} and diffusion methods~\cite{DBLP:conf/nips/DhariwalN21, DBLP:journals/corr/abs-2209-00796} over the past two years, there has been a notable proliferation of proposed change detection tasks. Thus, it is imperative to conduct a comprehensive change detection survey that encompasses the latest state-of-the-art methodologies.
%\xt{the contributions can be considered as scope or content since no new methods are proposed.}

  \par \noindent In summary, this survey has three main contributions.
  \begin{itemize}
      \item Compared to the previous survey works in change detection~\cite{DBLP:journals/tc/Lillestrand72, doi:10.1080/0143116031000139863, DBLP:journals/remotesensing/YouCZ20, DBLP:journals/ecoi/AfaqM21, DBLP:journals/remotesensing/JiangPZXHLMH22, doi:10.1080/10095020.2022.2128902, DBLP:journals/remotesensing/ShafiqueCKAA22, STILLA2023228, DBLP:journals/remotesensing/LiZGZX22}, our study provides a more comprehensive overview of the latest research on change detection in the past decade. Moreover, we systematically review the entirety of current algorithms to date and offer a relatively comprehensive depiction of the current state of the field. 
      \item Several systematical taxonomies of current change detection algorithms are presented from three perspectives: algorithm granularity, supervision modes, and learning frameworks. Furthermore, some influential works published with state-of-the-art performance on several dominant benchmarks are provided for future research.
      \item Change detection learning schemes and their applications remain a rapidly developing research field, as illustrated in Fig.~\ref{fig:paper_summary}, and many challenges need to be addressed. In this paper, we comprehensively review the existing challenges in this field and provide our perspective on future trends.
  \end{itemize}

\par \noindent \textbf{Scope.} The survey aims to cover the mainstream change detection algorithms from three perspectives: algorithm granularity, supervision modes, and learning frameworks, as depicted in Fig.~\ref{fig:attention_transformer}. The algorithms within the three taxonomies are carefully selected to ensure their orthogonality, thereby complementing each other and maintaining their distinctiveness. The survey focuses on the most representative works, despite the abundance of preprints or published works available. Additionally, we benchmark several predominant datasets and provide a detailed discussion of future research trends.

 \par \noindent \textbf{Organization.}  This paper is organized as follows.
  In Sec.~\ref{pre}, we review the preliminary knowledge of change detection, i.e. task definition, dominant datasets, evaluation metrics, and some transformer basics. 
  In Sec.~\ref{method}, we take a closer look at existing methods and classify them from three perspectives. 
  Sec.~\ref{benchmark} benchmarks the state-of-the-art performance on several popular change detection datasets. 
  Sec.~\ref{future} raises the future research trends and discussions. Finally, we conclude this survey in Sec.~\ref{conclusion}.
  \section{Preliminary Knowledge} % 2-3 pages
 % \xt{Could we shorten the dataset parts but add a review of non-deep learning methods for reference?}
  \label{pre}
  \par In this section, we will first introduce the task of change detection in remote sensing and provide a comprehensive definition. Subsequently, we will delve into various popular change detection datasets from diverse data sources and present some crucial evaluation metrics that will be utilized in the benchmark part to assess the performance of different methods. Finally, we will introduce some basic knowledge of transformers that are extensively utilized in modern attention and transformer-based algorithms.
 \begin{figure}
 \centering
  \includegraphics[width=0.5\textwidth]{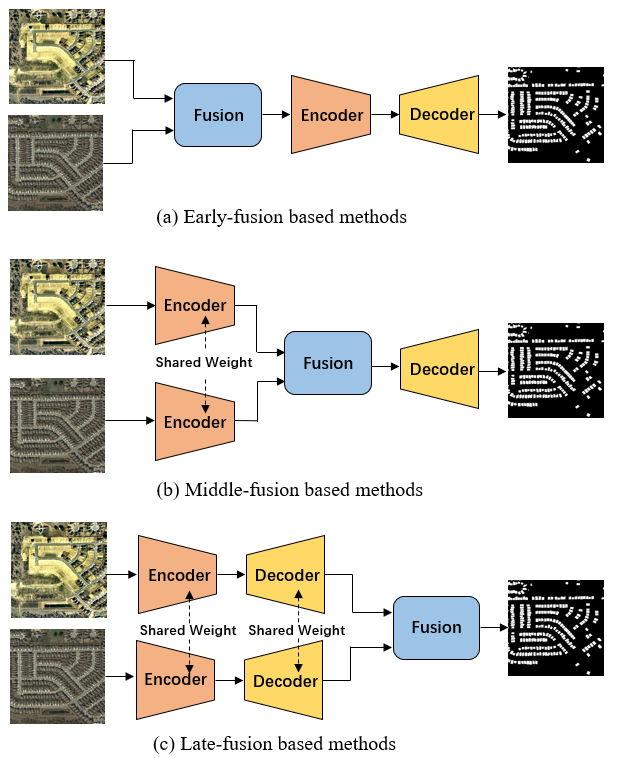}
  \caption{The existing frameworks for the change detection task. They can be classified into three categories based on the position of the fusion module, including early-fusion methods, middle-fusion methods, and late-fusion methods. Please note that this criterion is not limited to the deep learning methods, traditional methods also apply to it.}
 \label{fig:framework}
\end{figure}
  \subsection{Task Definition}
\par Change detection in remote sensing refers to the process of identifying and quantifying changes in the land surface over time using remotely sensed data. It can be formally defined as follows: As Fig.~\ref{fig:framework} shows, given two or more remotely sensed images of the same area captured at different times, the change detection task aims to identify changes in the land surface, such as changes in land cover or land use, natural phenomena, or human activities, by comparing the pixel values of the images and applying appropriate change detection algorithms. According to the position of the fusion module, existing change detection algorithms, especially the deep learning methods, can be categorized into three groups: early-fusion methods~(Fig.~\ref{fig:framework}(a)), middle-fusion methods~(Fig.~\ref{fig:framework}(b)), and late-fusion methods~(Fig.~\ref{fig:framework}(c)).

\par Moreover, the whole pipeline of the change detection task can be divided into several subtasks, including:
\begin{itemize}
    \item \textbf{Image preprocessing}: This involves removing noise, correcting geometric and radiometric distortions, and enhancing image quality to ensure the images are properly aligned and suitable for analysis.
    \item \textbf{Image registration}: This involves aligning the images spatially and temporally to ensure that corresponding pixels in each image are accurately compared.
    \item \textbf{Change detection algorithm selection}: To accurately detect changes in images, it is crucial to select the appropriate change detection algorithms based on the application and image characteristics. This can include traditional methods, as well as more advanced techniques such as CNN methods or transformer methods.
    \item \textbf{Post-processing}: This involves removing noise and false positives, and generating a final change map that accurately represents the changes. It is worth noting that this step is not mandatory and can be skipped.
\end{itemize}

\begin{table*}[!ht]
  \centering
  \caption{The summary of advantages and disadvantages for different data sources in remote sensing change detection task.}
  \scalebox{1.0}{
  \begin{tabular}
  {p{0.15\textwidth}|p{0.38\textwidth}|p{0.38\textwidth}}
    \hline\hline
  \rowcolor{gray!30!}  & \makecell[c]{\textbf{Advantages}} & \makecell[c]{\textbf{Disadvantages}} \\
    \hline
    \textbf{SAR Data} & \makecell[l]{
 1) Can penetrate through vegetation and clouds; \\
 2) Can detect subtle changes in object scattering; \\
 3) Can provide information on surface deformation; \\
 4) Can work well in all weather conditions.} & \makecell[l]{1) Can be susceptible to geometric distortion; \\ 2) Can be subject to electromagnetic interference; \\ 3) Can be complex and difficult to interpret; \\ 4) SAR sensors are expensive to develop and maintain.} \\
    \hline
    \textbf{Multi-spectral Data} & \makecell[l]{1) Can distinguish materials based on spectral information; \\ 2) Multi-spectral sensors are relatively inexpensive; \\ 3) Multi-spectral images are widely available. } & \makecell[l]{1) Multi-spectral sensors have limited spectral resolution; \\ 2) Images are susceptible to atmospheric interference; \\ 3) Images are affected by land cover and seasonal changes.}  \\
    \hline
    \textbf{Hyper-spectral Data} & \makecell[l]{1) Can distinguish materials with similar spectral signatures; \\ 2) Can provide rich information about the chemical and \\ physical properties of materials.} & \makecell[l]{1) Hyperspectral sensors are relatively expensive; \\ 2) Hyperspectral sensors may have limited spatial resolution; \\ 3) Images are susceptible to atmospheric interference.} \\
    \hline
    \textbf{Heterogeneous Data} & \makecell[l]{1) Heterogeneous images can provide complementary \\ information and improve the overall accuracy and quality of \\ the output by combining data from different sensors.} & \makecell[l]{1) The integration of heterogeneous images can be complex \\ and challenging; Poor quality or mismatched data can lead to \\ artifacts, noise, or errors in the fusion process; \\ }\\
    \hline
    \textbf{3D Change Detection Data} & \makecell[l]{1) Can capture fine details and subtle changes; \\ 2) Can penetrate through dense vegetation and provide \\ accurate elevation information.}  & \makecell[l]{1) Collection and processing are costly and time-consuming; \\ 2) The accuracy and quality of the output depend on the \\ quality of the calibration and the presence of noise.}\\
    \hline\hline
  \end{tabular}}
  \label{tab:data_advantage}
\end{table*}
\par In this paper, our focus lies primarily on the change detection algorithms. For a comprehensive understanding of the other steps involved, the readers are recommended to refer to existing works~\cite{doi:10.1080/0143116031000139863, DBLP:journals/tip/RadkeAAR05, DBLP:journals/ecoi/AfaqM21}.
\par As shown in Fig.~\ref{fig:framework}, given two input data pairs, $\textbf{D}_{1} \in \mathcal{R}^{P \times N}$ and $\textbf{D}_{2} \in \mathcal{R}^{P \times M}$, captured at the same location at different times. For images, $P$ denotes $H \times W$ for the images with width $W$ and height $H$. For point cloud data, $P$ represents the number of points. Specifically, the change detection algorithms can be summarized as follows:
\begin{equation}
    \textbf{M} = CD (PR(\textbf{D}_{1}), PR(\textbf{D}_{2})),
\end{equation}
where $PR$ denotes the image \textbf{p}reprocessing and \textbf{r}egistration steps to ensure that the paired images are well aligned. $CD$ denotes the change detection algorithm. The output maps of the changed area are represented by $\textbf{M}$. According to the different input data sources, both $\textbf{D}_{1}$ and $\textbf{D}_{2}$ can come from homogeneous data sources, such as Synthetic Aperture Radar (SAR) images, multi-spectral images (including the optical images), hyperspectral images and point cloud data. $\textbf{D}_{1}$ and $\textbf{D}_{2}$ can also originate from different data sources, which is referred to as ``heterogeneous data''. Based on different types of output, change detection tasks can be classified into two categories: binary change detection~\cite{DBLP:conf/igarss/AlvarezRD20, DBLP:journals/tgrs/LvLB20}, which differentiates between changed and unchanged areas, and semantic change detection~\cite{tian2020hi, DBLP:journals/tgrs/DingGLMZB22, peng2021scdnet}, which identifies changes in the categories or labels of objects before and after the change.
%The output of change detection is a map that shows the location and magnitude of changes, which can be used for a variety of applications, such as land management, environmental monitoring, and disaster response.

  \subsection{Change Detection Datasets} 
  \par In the following, we will summarize some dominant change detection datasets from the past few decades. To facilitate correspondence with various methods, the datasets are presented in five distinct categories based on the type of data sources, i.e. SAR data, multi-spectral data, hyperspectral data, heterogeneous data, and 3D change detection data. For more details about the change detection datasets, readers can refer to the project websites\footnote{\url{https://github.com/wenhwu/awesome-remote-sensing-change-detection}}\footnote{\url{https://github.com/yjt2018/awesome-remote-sensing-change-detection}}.

\begin{table*}[!ht]
  \scriptsize
  \centering
  \caption{The detailed information of several dominant change detection datasets. Note that the quantity of data is represented by the number of data pairs.}
  \scalebox{1.0}{
  \begin{tabular}
  {p{0.12\textwidth}|p{0.1\textwidth}|p{0.12\textwidth}|p{0.16\textwidth} |p{0.12\textwidth}|p{0.1\textwidth}|p{0.1\textwidth}}
    \hline\hline
  \rowcolor{gray!60!}  \makecell[c]{\textbf{Dataset}} & \makecell[c]{\textbf{Resolution}} & \makecell[c]{\textbf{quantity of Data}} & \makecell[c]{\textbf{Location}} & \makecell[c]{\textbf{Satellite Types}} & \makecell[c]{\textbf{Capture Date}} & \makecell[c]{\textbf{Categories}}\\
  \hline \hline
  \rowcolor{gray!30!} \multicolumn{7}{c} {\textbf{SAR Dataset}}  \\
    \hline
    \makecell[c]{Yellow River~\cite{9982693}} & \makecell[c]{$7666\times 7692$} & \makecell[c]{1} & \makecell[c]{Yellow River Estuary, \\ China} & \makecell[c]{Radarsat-2} & \makecell[c]{June 2008 \\ June 2009} & \makecell[c]{2} \\
    \hline
    \makecell[c]{Bern~\cite{9829884}} & \makecell[c]{$301\times 301$} & \makecell[c]{1} & \makecell[c]{Bern, Switzerland} & \makecell[c]{European Remote \\ Sensing Satellite-2} & \makecell[c]{April 1999 \\ May 1999} & \makecell[c]{2} \\
    \hline\hline
  \rowcolor{gray!30!}  \multicolumn{7}{c} {\textbf{Multi-spectral Dataset}}  \\
    \hline
    \makecell[c]{LEVIR-CD~\cite{Chen2020}} & \makecell[c]{$1024\times 1024$ \\ 0.5 m/pixel} & \makecell[c]{637} & \makecell[c]{American cities} & \makecell[c]{Google Earth} & \makecell[c]{2002 to 2018} & \makecell[c]{2} \\
    \hline
    \makecell[c]{CDD~\cite{Lebedev2018CHANGEDI}} & \makecell[c]{$4725\times 2700$ \\ $1900 \times 1000$ \\ 0.03-1 m/pixel} & \makecell[c]{7 \\ 4} & \makecell[c]{--} & \makecell[c]{Google Earth} & \makecell[c]{--} & \makecell[c]{2} \\
    \hline
    \makecell[c]{WHU Building~\cite{8444434}} & \makecell[c]{$512 \times 512$\\ 0.075 m/pixel \\ 0.3-2.5 m/pixel \\ 2.7 m/pixel} & \makecell[c]{8189 \\ 102 \\ 17388} & \makecell[c]{Christchurch, New Zealand \\ Cities over the world \\ East Asia} & \makecell[c]{QuickBird, \\ Worldview series, \\ IKONOS, and ZY-3} & \makecell[c]{--} & \makecell[c]{2} \\
    \hline
    \makecell[c]{SECOND~\cite{DBLP:journals/tgrs/ZhangJLYSL22}} & \makecell[c]{$512 \times 512$ \\ 0.5-3 m/pixel} & \makecell[c]{4662} & \makecell[c]{Hangzhou, Chengdu, \\ Shanghai, China} & \makecell[c]{--} & \makecell[c]{--} & \makecell[c]{6} \\
    \hline\hline
  \rowcolor{gray!30!}  \multicolumn{7}{c} {\textbf{Hyperspectral Dataset}}  \\
    \hline
    \makecell[c]{River~\cite{Wang2019GETNETAG}} & \makecell[c]{$463 \times 241$} & \makecell[c]{1} & \makecell[c]{Jiangsu, China} & \makecell[c]{EO-1 sensor} & \makecell[c]{May 3, 2013 \\ Dec. 21, 2013} & \makecell[c]{2}\\
    \hline
    \makecell[c]{Hermiston~\cite{DBLP:journals/tgrs/WangWWWB22}} & \makecell[c]{$307 \times 241$} & \makecell[c]{1} & \makecell[c]{Hermiston city, USA} & \makecell[c]{Hyperion sensor} & \makecell[c]{May 1, 2004 \\ May 8, 2007} & \makecell[c]{2} \\
    \hline\hline
  \rowcolor{gray!30!}  \multicolumn{7}{c} {\textbf{Heterogeneous Dataset}}  \\
    \hline
    \makecell[c]{California~\cite{9773305}} & \makecell[c]{$850 \times 500$} & \makecell[c]{one multi-spectral \\ one SAR} & \makecell[c]{California} & \makecell[c]{Landsat-8\\ Sentinel-1A} & \makecell[c]{ Jan. 5, 2017 \\ Feb. 18, 2017} & \makecell[c]{2} \\
    \hline\hline
      \rowcolor{gray!30!}  \multicolumn{7}{c} {\textbf{3D Change Detection Dataset}}  \\
    \hline
    \makecell[c]{3DCD~\cite{isprs-archives-XLIII-B3-2022-1349-2022}} & \makecell[c]{$400 \times 400$ \\ 0.5 m/pixel \\ $200 \times 200 $ \\ 1 m/pixel} & \makecell[c]{472 2D \\  472 3D} & \makecell[c]{Valladolid, Spain} & \makecell[c]{--} & \makecell[c]{2010 \\ 2017} & \makecell[c]{2} \\
    \hline\hline  
\end{tabular}}
  \label{tab:dataset_summary}
\end{table*}

\par \noindent \textbf{SAR Data.} Synthetic Aperture Radar (SAR), as a special type of radar, utilizes electromagnetic signals to generate two-dimensional or three-dimensional images of objects. These images can be applied for multiple purposes, such as object detection, geographical localization, and geophysical property estimation of complex environments. SAR has the advantage of working in all weather conditions, penetrating through vegetation and clouds, and has high-resolution imaging capabilities to identify small features, making it suitable for remote sensing analysis. However, SAR systems can be susceptible to geometric distortion, electromagnetic interference, and speckle noise, which need to be addressed in industrial applications (shown in the first block of Tab.~\ref{tab:data_advantage}). For the SAR datasets, we present two dominant datasets, i.e. Yellow River~\cite{9982693} and Bern~\cite{9829884}, along with their specific details shown in Tab.~\ref{tab:dataset_summary}.
\par \noindent \textbf{Multi-spectral Data.} Multi-spectral images can be used to identify and map various features on the Earth's surface, such as vegetation, water bodies, and urban areas. Each type of feature reflects or emits energy uniquely, and the information captured by the different spectral bands can be used to distinguish between them. For example, healthy vegetation typically reflects more energy in the near-infrared part of the spectrum than other types of land cover, which makes it possible to identify and map areas of vegetation using multi-spectral data. The advantages and disadvantages of multi-spectral images are detailed in the second block of Tab.~\ref{tab:data_advantage}. Four primary multi-spectral datasets, i.e. LEVIR-CD~\cite{Chen2020}, CDD~\cite{DBLP:journals/corr/abs-2108-07955}, WHU Building~\cite{8444434}, and SECOND~\cite{DBLP:journals/tgrs/ZhangJLYSL22}, are introduced.
\par \noindent \textbf{Hyperspectral Data.} Hyperspectral imaging is a technique that captures and analyzes numerous narrow, contiguous spectral bands across the electromagnetic spectrum. Unlike multi-spectral images, which typically capture data in a few broad spectral bands, hyperspectral images capture data across hundreds of spectral bands, providing highly detailed information about the composition and characteristics of the image materials. The high spectral resolution of hyperspectral images enables the identification and discrimination of materials that may have similar appearances but different spectral signatures, such as different types of vegetation or minerals. Hyperspectral imaging is widely used in remote sensing applications, such as environmental monitoring, mineral exploration, and crop health assessment. The advantages and disadvantages of hyperspectral images are detailed in the third block of Tab.~\ref{tab:data_advantage}. For the hyperspectral data, River~\cite{Wang2019GETNETAG} and Hermiston~\cite{DBLP:journals/tgrs/WangWWWB22} are utilized in this survey. The details of these datasets can be found in Tab.~\ref{tab:dataset_summary}.
\par \noindent \textbf{Heterogeneous Data}. Heterogeneous data in remote sensing refers to images that combine data from multiple sources or sensors with different characteristics, such as SAR images, multi-spectral images, hyperspectral images, and LiDAR data. By integrating data from different sources, heterogeneous images can provide more comprehensive and accurate information about the Earth's surface than individual sensor data alone. For example, combining optical and SAR data can enable better identification and characterization of land cover types, while integrating LiDAR data can provide information about the three-dimensional structure of vegetation and terrain. Heterogeneous image analysis techniques have become increasingly important in remote sensing applications such as land use and land cover mapping, environmental monitoring, and disaster management. However, the integration and processing of data from multiple sources also pose significant challenges, including data registration, normalization, and fusion. The advantages and disadvantages of heterogeneous images are summarized in the fourth block of Tab.~\ref{tab:data_advantage}. In this survey, we introduce the California dataset~\cite{DBLP:journals/tnn/YangJLHYJ22}, which includes both SAR images and multi-spectral images. A summary of this dataset is provided in Tab.~\ref{tab:dataset_summary}.
\par \noindent \textbf{3D Change Detection Data.} 3D point cloud data refers to a collection of points in three-dimensional space that represent the surface of an object or a terrain. In remote sensing, 3D point cloud data is obtained by using Light Detection and Ranging (LiDAR) technology, which uses laser pulses to measure the distance between the sensor and the ground or other objects. For change detection applications, 3D change detection data can be used to identify differences in terrain or object heights and shapes over time. This is particularly useful for monitoring natural and man-made features such as buildings, vegetation, and coastlines. The advantages and disadvantages of 3D change detection data are summarized in the final row of Tab.~\ref{tab:data_advantage}. For the 3D change detection dataset, 3DCD~\cite{isprs-archives-XLIII-B3-2022-1349-2022} is employed to benchmark the state-of-the-art performance in this survey.

\subsection{Evaluation Metrics}
\par Evaluation metrics are essential in evaluating the performance of a change detection model. 
This study presents a concise introduction and analysis of common evaluation metrics. These include true positive (TP) and true negative (TN), which indicate the number of correctly identified changed and unchanged pixels, respectively. 
On the other hand, false positives (FP) and false negatives (FN) refer to the number of pixels wrongly classified as changed or unchanged, respectively. 
A high precision value implies the algorithm's ability to identify changed pixels accurately. 
In contrast, a high recall value suggests the algorithm can detect a higher proportion of changed pixels from the ground truth data. 
The Intersection over Union (IoU) represents the ratio of intersection and concatenation of the predicted map and ground truth. 
The overall accuracy (OA) metric indicates the prediction's accuracy. The F1-score is a harmonic average of precision and recall. 
Additionally, other metrics are designed for specific applications, such as the Kappa coefficient (KC)~\cite{Rosenfield1986ACO}, which measures classification accuracy based on the confusion matrix. 
RMSE (Root Mean Squared Error) and cRMSE (Changed Root Mean Squared Error)~\cite{MARSOCCI2023325} are commonly used for evaluating 3D change detection model change detection performance. 
cRMSE considers only errors in ground truth pixels influenced by altitude. In semantic change detection (SCD), mean Intersection over Union (mIoU), Separated Kappa (Sek), and SCD-targeted F1 Score ($F_{scd}$) are used to evaluate accuracy. 
For more information about these metrics, readers can refer to~\cite{DBLP:journals/tgrs/DingGLMZB22}.

\subsection{Transformer Basics}
 \begin{figure}
 \centering
  \includegraphics[width=0.48\textwidth]{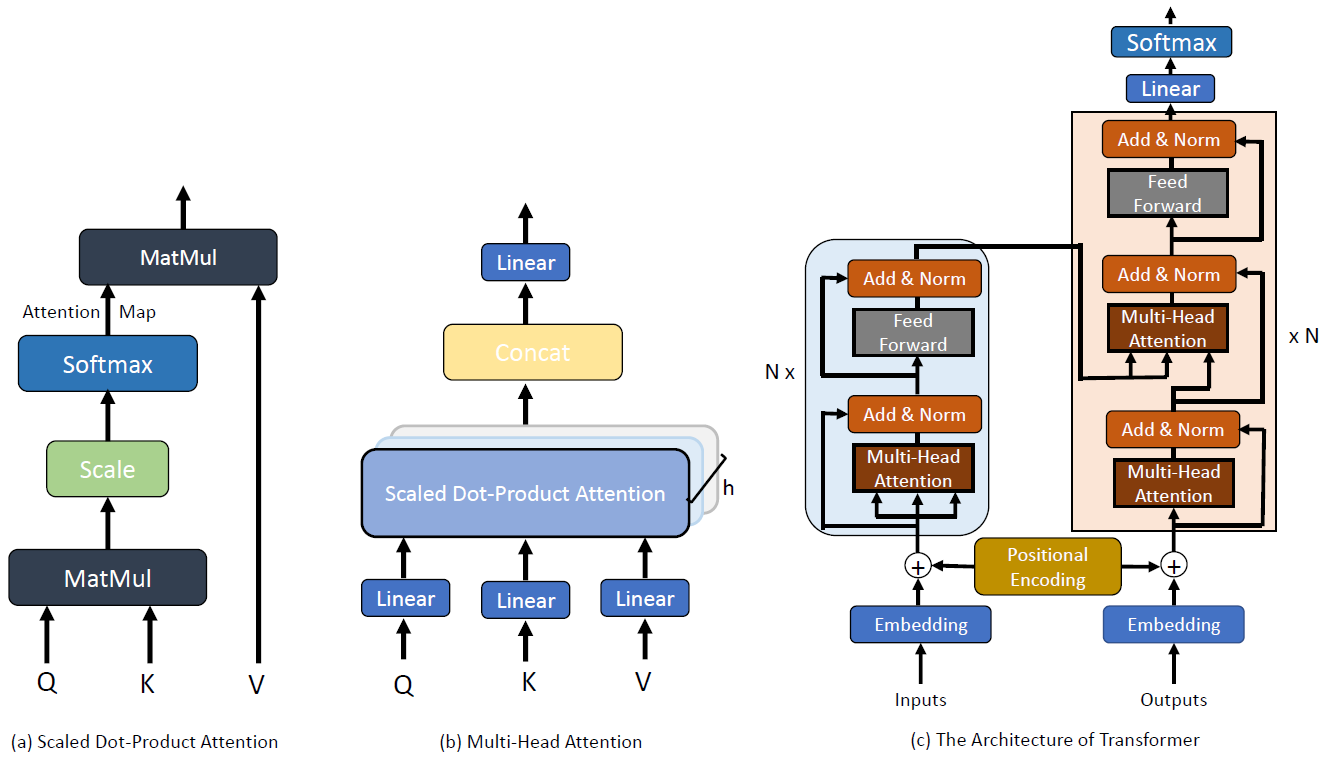}
  \caption{Some preliminary knowledge of transformer.}
 \label{fig:transformer}
\end{figure}
\par \noindent \textbf{Attention Basics:} Attention~\cite{DBLP:conf/nips/VaswaniSPUJGKP17} algorithms are a type of machine learning algorithm that selectively focus on certain aspects of data or inputs while ignoring others. They assign different weights or importance values to different parts of the input, based on their relevance to the task at hand. Attention algorithms have become increasingly popular in natural language processing (NLP)~\cite{DBLP:journals/tnn/GalassiLT21} and computer vision applications~\cite{DBLP:journals/cvm/GuoXLLJMZMCH22}, as they enable models to focus on the most relevant information in a sentence or image. As depicted in Fig.~\ref{fig:transformer} (a), given the Key ($K$), Query ($Q$), and Value ($V$), the attention mechanism can be defined as:
\begin{equation}
    \text{Att}(Q,K,V) = \text{Softmax}(\frac{QK^{T}}{\sqrt{d_{k}}})V.
    \label{equ:attention}
\end{equation}
The scale factor $d_k$ represents the dimension of the keys in the attention mechanism. The resulting attention map is multiplied with $V$ to focus on the most relevant regions. In practice, a multi-head attention (MHA) module with $h$ heads is often used instead of a single attention function shown in Eq.~\ref{equ:attention}. As shown in Fig.~\ref{fig:transformer} (b), the MHA involves performing multiple attention operations in parallel and concatenating the attention results. The concatenated outputs are then fused together using a projection matrix $W_{P}$, which allows the model to effectively incorporate information from multiple heads of attention.
\begin{equation}
    \text{MHA}(Q,K,V) = W_{P} \text{Concat}([Att_1, Att_2, ..., Att_h]).
    \label{eq:MHA}
\end{equation}
Based on the different sources of Query ($Q$) and Key ($K$), MHA can be divided into multi-head self-attention (MHSA) and multi-head cross-attention (MHCA). MHSA refers to the case where the $Q$ and $K$ for all heads come from the same sequences or tokens, while MHCA refers to the case where $Q$ and $K$ for different heads come from different sequences or tokens.

\par \noindent \textbf{Transformer Basics:} As shown in Fig.~\ref{fig:transformer} (c),  the Transformer architecture~\cite{Transformer_survey} comprises an encoder and a decoder, both of which contain multiple layers of Multi-Head Attention (MHA) and feedforward neural networks (FFN). 
In particular, the FFN consists of two consecutive Multi-Layer Perceptrons (MLPs) with a non-linear activation function. 
In addition to the MHSA, the decoder utilizes MHCA to attend to the relevant information or regions from different multi-modal sequences. 
Unlike the CNNs and attention mechanism, the transformer adds positional encoding information to input and output embedding to capture the positional information.
  \section{Methodology: A Survey} % 3 pages
  \label{method}
 \begin{table*}[!ht]
  \scriptsize
  \centering
  \caption{Representative literature on various algorithm granularities for different data sources in remote sensing change detection tasks. We also illustrate the fusion category for each corresponding approach.}
  \scalebox{1.0}{
  \begin{tabular}
  {p{0.13\textwidth}|p{0.12\textwidth}|p{0.08\textwidth} |p{0.05\textwidth}|p{0.5\textwidth}}
    %\toprule %{c|l|m{7cm}}
    \hline\hline
    %\hline
  \rowcolor{gray!60!}  \makecell[c]{\textbf{Method}} & \makecell[c]{\textbf{Source}} & \makecell[c]{\textbf{Category}} & \makecell[c]{\textbf{Fusion}} & \makecell[c]{\textbf{Highlights}}\\
  \hline \hline
  \rowcolor{gray!30!}  \multicolumn{5}{c} {\textbf{SAR Methods}}  \\
    \hline
    \makecell[c]{ARCT+ \cite{DBLP:journals/lgrs/AtaseverG22}} & \makecell[c]{GRSL 2022} & \makecell[c]{Pixel-based} & \makecell[c]{Middile} & \makecell[l]{The arc-tangential subtraction operator is applied to obtain a difference image, which \\ is then subjected to K-means++ clustering to identify the changed regions.} \\
    \hline
    \makecell[c]{JR-KSVD \cite{9810349}} & \makecell[c]{J-STARS 2022} & \makecell[c]{Pixel-based} & \makecell[c]{Late}& \makecell[l]{ proposes a joint-related dictionary learning algorithm based on K-SVD, and an iterative \\ adaptive threshold optimization.} \\
    \hline
    \makecell[c]{Incoherent CDA~\cite{DBLP:journals/lgrs/VinholiSMP22}} & \makecell[c]{GRSL 2022} & \makecell[c]{Pixel-based} & \makecell[c]{Early}& \makecell[l]{utilizes a segmentation CNN to localize potential changes, and a classification CNN to \\ further inspect potential changes as true changes or false alarms.} \\
    \hline
    \makecell[c]{WMMs~\cite{DBLP:journals/tgrs/YangYYSX16}} & \makecell[c]{TGRS 2016} & \makecell[c]{Region-based} & \makecell[c]{Middle}& \makecell[l]{segments the PolSAR images into compact local regions, and then wishart mixture \\ models (WMMs) are used to model each local region.} \\
    \hline
    \makecell[c]{OBIA~\cite{DBLP:journals/tgrs/AmitranoGI21}} & \makecell[c]{TGRS 2022} & \makecell[c]{Region-based} & \makecell[c]{Middle}& \makecell[l]{takes advantages of consolidated SAR techniques and modern geographical object-based \\ image analysis (GEOBIA).} \\
    \hline
    \makecell[c]{UAFS-HCD~\cite{DBLP:journals/staeors/LuLCZXK15}} & \makecell[c]{J-STARS 2015} & \makecell[c]{Hybrid} & \makecell[c]{Late}& \makecell[l]{gets the preliminary change mask with pixel-based change detection method and obtains \\ the final change mask using the object-based change detection method.} \\
    \hline
    \makecell[c]{DSF~\cite{DBLP:journals/remotesensing/JavedJLH20}} & \makecell[c]{Remote \\ Sensing 2020} & \makecell[c]{Hybrid} & \makecell[c]{Middle}& \makecell[l]{detects the change detection results by extending pixel-based object detection method \\ into an OBCD through the Dempster-Shafer theory.} \\
    \hline\hline
  \rowcolor{gray!30!}  \multicolumn{5}{c} {\textbf{Multi-spectral Methods}}  \\
    \hline
    \makecell[c]{ADHR-CDNet~\cite{DBLP:journals/tgrs/ZhangTXYLYXJZ22}} & \makecell[c]{TGRS 2022} & \makecell[c]{Pixel-based} & \makecell[c]{Early}& \makecell[l]{proposes an HRNet with differential pyramid module, and multiscale spatial feature \\ attention module is presented to fuse different information.} \\
    %\hline
    %\makecell[c]{Mask-CDNet~\cite{DBLP:journals/ijon/BuLHLL20}} & \makecell[c]{Neurocomputing \\ 2020} & \makecell[c]{Pixel-based} & \makecell[c]{\textbf{Late}}& \makecell[l]{proposes an end-to-end deep learning based module with mask proposal module and \\ mask fine-tune module.} \\
    \hline
    \makecell[c]{LWCDNet~\cite{DBLP:journals/lgrs/HanLZ22}} & \makecell[c]{GRSL 2022} & \makecell[c]{Pixel-based} & \makecell[c]{Early}& \makecell[l]{proposes a lightweight fully convolution network with convolutional block attention \\ module and Lov-wce loss.} \\
    \hline
    \makecell[c]{COCRF~\cite{DBLP:journals/tgrs/ShiZZLLZ22}} & \makecell[c]{TGRS 2022} & \makecell[c]{Region-based} & \makecell[c]{Early}& \makecell[l]{proposes a  class-prior object-oriented conditional random field framework to handle \\ binary and multiclass change detection tasks.} \\
    \hline
    \makecell[c]{SSS-CD~\cite{DBLP:journals/remotesensing/GeDMHP22}} & \makecell[c]{Remote Sensing \\ 2022} & \makecell[c]{Region-based} & \makecell[c]{Early}& \makecell[l]{integrates spectral–spatial–saliency change information and fuzzy integral decision \\ fusion for the change detection task.} \\
    \hline
    \makecell[c]{DP-CD-Net~\cite{DBLP:journals/lgrs/JiangXWT22}} & \makecell[c]{GRSL 2022} & \makecell[c]{Hybrid} & \makecell[c]{Early}& \makecell[l]{proposes a dual-pathway feature difference network, an adaptive fusion module, and \\ an auxiliary supervision strategy.} \\
    \hline\hline
  \rowcolor{gray!30!}  \multicolumn{5}{c} {\textbf{Hyperspectral Methods}}  \\
    \hline
    \makecell[c]{GETNET~\cite{Wang2019GETNETAG}} & \makecell[c]{TGRS 2019} & \makecell[c]{Pixel-based} & \makecell[c]{Early}& \makecell[l]{proposes a general end-to-end 2D CNN for hyperspectral \\ image change detection.} \\
    \hline
    \makecell[c]{SSA-SiamNet~\cite{9494085}} & \makecell[c]{TGRS 2022} & \makecell[c]{Pixel-based} & \makecell[c]{Middle}& \makecell[l]{proposes an end-to-end siamese CNN with a spectral-spatial-wise attention mechanism.} \\
    \hline
    \makecell[c]{CDFormer~\cite{DBLP:journals/lgrs/DingLZ22}} & \makecell[c]{GRSL 2022} & \makecell[c]{Pixel-based} & \makecell[c]{Middle}& \makecell[l]{introduces a transformer encoder to the hyperspectral image change detection framework.} \\
    \hline
    \makecell[c]{FuzCVA~\cite{DBLP:conf/igarss/Erturk18}} & \makecell[c]{IGARSS 2018} & \makecell[c]{Hybrid} & \makecell[c]{Middle}& \makecell[l]{proposes a fuzzy inference combination strategy that combines the angle and \\ magnitude distances.} \\
    \hline
    \makecell[c]{MSDFFN~\cite{DBLP:journals/tgrs/LuoZLGGR23}} & \makecell[c]{TGRS 2023} & \makecell[c]{Hybrid} & \makecell[c]{Middle}& \makecell[l]{proposes bidirectional diff-changed feature representation module and a multiscale \\ attention fusion module to fuse the changed features.} \\
    \hline\hline
  \rowcolor{gray!30!}  \multicolumn{5}{c} {\textbf{Heterogeneous Methods}}  \\
    \hline
    \makecell[c]{SCCN~\cite{DBLP:journals/tnn/LiuGQZ18}} & \makecell[c]{TNNLS 2018} & \makecell[c]{Pixel-based} & \makecell[c]{Middle}& \makecell[l]{proposes a symmetric convolutional
coupling network for unsupervised change \\ detection tasks.} \\  
    \hline
    \makecell[c]{SSPCN~\cite{DBLP:journals/staeors/LiGZW21}} & \makecell[c]{J-STARS 2021} & \makecell[c]{Pixel-based} & \makecell[c]{Middle}& \makecell[l]{introduces a classification method to get the pseudo labels, and then spatially \\ self-paced convolutional network to update the pseudo label labels to get better results.} \\
     \hline
    \makecell[c]{MSGCN~\cite{DBLP:journals/aeog/WuLQNZFS21}} & \makecell[c]{IJAEOG 2022} & \makecell[c]{Region-based} & \makecell[c]{Middle}& \makecell[l]{introduces a new change detection method based on the graph convolutional network \\ and multiscale object techniques.} \\
    \hline
    \makecell[c]{CMS-HCC~\cite{DBLP:journals/tgrs/WanXY19}} & \makecell[c]{TGRS 2019} & \makecell[c]{Region-based} & \makecell[c]{Middle}& \makecell[l]{proposes a region-based change detection method with a cooperative multitemporal \\ segmentation process and a hierarchical compound classification process.} \\
    \hline
    \makecell[c]{HMCNet~\cite{DBLP:journals/tgrs/WangL22b}} & \makecell[c]{TGRS 2022} & \makecell[c]{Hybrid} & \makecell[c]{Middle}& \makecell[l]{proposes an MLP-CNN hybrid model with multilayer perceptron and convolutional \\ neural network to achieve change detection result.} \\
    \hline
    \makecell[c]{CD-GAN~\cite{DBLP:journals/corr/abs-2203-00948}} & \makecell[c]{Arxiv 2022} & \makecell[c]{Hybrid} & \makecell[c]{Middle}& \makecell[l]{introduce a robust fusion-based adversarial framework that fuses the results \\ from predefined and previously trained networks.} \\
    \hline\hline
      \rowcolor{gray!30!}  \multicolumn{5}{c} {\textbf{3D Change Detection Methods}}  \\
    \hline
    \makecell[c]{HDG-nDSM~\cite{rs15051414}} & \makecell[c]{Remote \\ Sensing 2023} & \makecell[c]{Pixel-based} & \makecell[c]{Middle}& \makecell[l]{proposes a height difference-generated nDSM, including morphological filters and \\ criteria considering area size and shape parameters.} \\
    \hline
    \makecell[c]{DALE-CD~\cite{isprs-archives-XLIII-B2-2022-523-2022}} & \makecell[c]{ISPRS \\ Archives 2022} & \makecell[c]{Pixel-based} & \makecell[c]{Early}& \makecell[l]{proposes a 3D change detection method based on density adaptive local Euclidean \\ distance.} \\
    \hline
    \makecell[c]{CHM-CD~\cite{DBLP:journals/tgrs/MarinelliPB18}} & \makecell[c]{TGRS 2018} & \makecell[c]{Region-based} & \makecell[c]{Middle}& \makecell[l]{first detects the large changes, and then focuses on the individual-tree canopy to detect \\ the single-tree changes by mean of an object-based CD.} \\
    \hline\hline
  \end{tabular}}
  \label{tab:data_data_source}
\end{table*}
\par In this section, we will comprehensively review methodologies from three distinct perspectives: 1) Taxonomy based on algorithm granularity, 2) Taxonomy based on supervision modes, and 3) Taxonomy based on learning frameworks. It is imperative to note that the presented algorithms from these three taxonomies are carefully chosen to ensure orthogonality, thereby complementing one another. Furthermore, given the discrepancies in data quantities and unique data characteristics across various data sources, we will present a comprehensive introduction to the existing approaches categorized by data type in the following.
\subsection{Taxonomy based on Algorithm Granularity}
\par As introduced in the previous section, the remote sensing community categorizes data for the change detection task based on the types of available data sources. For each data source, we classify the change detection methods based on the algorithm granularity into the following categories: pixel-based methods, region-based methods, and hybrid methods.
\begin{itemize}
    \item \textbf{Pixel-based methods:} Pixel-based methods are commonly used for image segmentation tasks~\cite{long2015fully, li2020improving, DBLP:conf/cvpr/LiHLLCSWTL21, he2021boundarysqueeze} in computer vision. These methods assign a label to each individual pixel in an image based on its spectral characteristics, with the goal of partitioning the image into regions of different classes. Traditional pixel-based methods often suffer from false positives and false negatives. Fortunately, with the advent of deep learning and its increased receptive field, such as pyramid pooling module~\cite{DBLP:conf/cvpr/ZhaoSQWJ17}, atrous convolution~\cite{DBLP:journals/corr/ChenPSA17} and attention module~\cite{DBLP:conf/nips/VaswaniSPUJGKP17}, pixel classification methods based on deep learning can achieve significantly improved performance. It is worth noting that most end-to-end CNN models fall under the category of pixel-based methods.
    \item \textbf{Region-based methods:} Region-based methods~\cite{DBLP:journals/tgrs/YangYYSX16, DBLP:journals/remotesensing/ChenLZCYLZ22}, also known as object-based methods, leverage image segmentation techniques first to group pixels into meaningful regions, such as objects, superpixels, or bounding boxes, based on their spatial, spectral, and contextual characteristics. These grouped regions are then used as the units for detecting and recognizing the changed results with either traditional or deep learning methods. 
%\par This approach is particularly effective in reducing noise and detecting large changes that occur in objects with similar spectral characteristics, which helps to reduce false positives. However, it is computationally more intensive and may require manual intervention for object selection and segmentation. Moreover, the effectiveness of such methods ultimately depends on the quality of the grouped regions. 
    \item \textbf{Hybrid methods:} The use of hybrid methods~\cite{DBLP:journals/lgrs/LiLZSA015, DBLP:journals/tgrs/WangL22b, DBLP:journals/lgrs/CarvalhoJASB22} has been identified as a powerful approach for change detection tasks. These methods leverage the advantages of multiple individual techniques such as pixel-based methods, region-based methods, or a combination of both to achieve improved accuracy in change detection. By integrating different methodologies, hybrid approaches can address the limitations of each individual method and provide a more robust and comprehensive solution for detecting changes in remote sensing imagery. These methods typically involve the parallel or successive use of pixel-based and region-based techniques to detect changes from different perspectives. 
\end{itemize}
\par In the following, we will present an overview of change detection algorithms based on algorithm granularity for each available data source.

 \par \noindent \textbf{1) SAR Methods.} The first block in Tab.~\ref{tab:data_data_source} presents some representative literature for SAR change detection algorithms. Among the \textit{pixel-based methods}, Atasever \textit{et al.}~\cite{DBLP:journals/lgrs/AtaseverG22} propose an unsupervised change detection approach based on arc-tangential difference, gaussian and median filters and K-means++ clustering. JR-KSVD~\cite{9810349} introduces a joint-related dictionary learning algorithm based on the k-singular value decomposition and an iterative adaptive threshold optimization algorithm for unsupervised change detection. Vinholi \textit{et al.}~\cite{DBLP:journals/lgrs/VinholiSMP22} present an incoherent change detection algorithm based on CNNs, which includes a segmentation CNN to localize potential changes, and a classification CNN to further analyze the potential changes to classify them as real changes or false alarms. Among the \textit{region-based methods}, Yang \textit{et al.}~\cite{DBLP:journals/tgrs/YangYYSX16} propose a region-based change detection method that first utilizes the customized simple-linear-iterative-clustering algorithm~\cite{DBLP:journals/pami/AchantaSSLFS12} to generate superpixels, and then employs Wishart mixture models to model each local superpixel. Amitrano \textit{et al.}~\cite{DBLP:journals/tgrs/AmitranoGI21} exploit multitemporal geographical object-based image analysis, which involves classifying the data using a dictionary representation approach and combining class-specific information layers through fuzzy logic to bring out the underlying meaning in the data. For the \textit{hybrid methods}, UAFS-HCD~\cite{DBLP:journals/staeors/LuLCZXK15} proposes a new unsupervised fusion method for change detection, which employs an intuitive decision-level fusion scheme of pixel-based method and region-based method. Javed \textit{et al.}~\cite{DBLP:journals/remotesensing/JavedJLH20} propose a new approach for detecting recently developed urban areas by utilizing the Dempster-Shafer theory to extend the results of the pixel-based method into a region-based method. 
 
 \par \noindent \textbf{2) Multi-spectral Methods.} The second block of Tab.~\ref{tab:data_data_source} showcases a collection of representative literature on methods for change detection in multi-spectral images. Among the \textit{pixel-based methods}, ADHR-CDNet~\cite{DBLP:journals/tgrs/ZhangTXYLYXJZ22} proposes an attentive differential high-resolution change detection network, which introduces a new high-resolution backbone with a differential pyramid module. Bu \textit{et al.}~\cite{DBLP:journals/ijon/BuLHLL20} propose a new deep learning framework for change detection tasks, which consists of two collaborative modules to improve the estimation accuracy and computation efficiency. LWCDNet~\cite{DBLP:journals/lgrs/HanLZ22} proposes a lightweight convolution network for change detection with a typical encoder-decoder structure, an artificial padding convolution module, and a convolutional block attention module. For the \textit{region-based methods}, Shi \textit{et al.}~\cite{DBLP:journals/tgrs/ShiZZLLZ22} introduce a class-prior object-oriented conditional random field framework, which consists of a binary change detection task and a multiclass change detection task. Ge \textit{et al.}~\cite{DBLP:journals/remotesensing/GeDMHP22} propose an object-oriented change detection approach that integrates spectral-spatial-saliency change information and fuzzy integral decision fusion to eliminate the detection noise. Among the \textit{hybrid methods}, DP-CD-Net~\cite{DBLP:journals/lgrs/JiangXWT22} introduces a dual-pathway change detection network that compromises a dual-pathway feature difference network, an adaptive fusion module, and an auxiliary supervision strategy. Different from SAR images, multi-spectral images, especially very high-resolution images, tend to have a higher spatial resolution, thus some region-based, object-based, and superpixel-based approaches have been proposed to achieve both homogeneous results and efficiency. To delve deeper into this topic, readers can refer to the survey paper~\cite{DBLP:journals/remotesensing/YouCZ20} for additional details.
 
  \par \noindent \textbf{3) Hyperspectral Methods.} Some representative literature of hyperspectral image change detection is shown in the third block of Tab.~\ref{tab:data_data_source}. Among the \textit{pixel-based methods}, GETNET~\cite{Wang2019GETNETAG} presents a general end-to-end 2D CNN framework for hyperspectral image change detection. Wang \textit{et al.}~\cite{9494085} propose an end-to-end Siamese CNN with a spectral-spatial-wise attention mechanism to emphasize informative channels and locations, and suppress less informative ones to refine the spectral-spatial features adaptively. CDFormer~\cite{DBLP:journals/lgrs/DingLZ22} introduces a transformer encoder method for the hyperspectral image change detection tasks with space and time encodings, as well as a self-attention component. For the \textit{hybrid methods}, FuzCVA~\cite{DBLP:conf/igarss/Erturk18} presents a new fuzzy inference combination strategy to combine the angle and magnitude distances, which can provide improved change detection performance. Luo \textit{et al.}~\cite{DBLP:journals/tgrs/LuoZLGGR23} propose a multiscale diff-changed feature fusion network, which combines a temporal feature encoder-decoder subnetwork and a cross-layer attention module.
  \par \noindent \textbf{4) Heterogeneous Methods.} The fourth block of Tab.~\ref{tab:data_data_source} displays a selection of notable literature on techniques for detecting changes in heterogeneous data. Among the \textit{pixel-based methods}, SCCN~\cite{DBLP:journals/tnn/LiuGQZ18} presents a symmetric convolutional coupling network with a symmetrical structure and feature transformation. Li \textit{et al.}~\cite{DBLP:journals/staeors/LiGZW21} introduce a spatially self-paced convolutional network for change detection in an unsupervised way. For the \textit{region-based methods}, MSGCN~\cite{DBLP:journals/aeog/WuLQNZFS21} proposes a new change detection method that combines graph convolutional network and multiscale object-based technique for both homogeneous and heterogeneous images. Wan \textit{et al.}~\cite{DBLP:journals/tgrs/WanXY19} propose an improved change detection method that combines a cooperative multitemporal segmentation method and a region-based multitemporal hierarchical Markov random field model. For the \textit{hybrid methods},  HMCNet~\cite{DBLP:journals/tgrs/WangL22b} introduces multilayer perceptron into CNNs for change detection tasks. Wang \textit{et al.}~\cite{DBLP:journals/corr/abs-2203-00948} introduce a deep adversarial network to fuse a pair of multiband images, which can be easily complemented by a network with the same architecture to perform change detection.
  
 \par \noindent \textbf{5) 3D Change Detection Methods.} The final block of Tab.~\ref{tab:data_data_source} demonstrates some representative literature for 3D change detection tasks. Among the \textit{pixel-based methods}, Marmol \textit{et al.}~\cite{rs15051414} propose an algorithm that is based on height difference-generated nDSM, including morphological filters and criteria considering area size and shape parameters. Chai \textit{et al.}~\cite{isprs-archives-XLIII-B2-2022-523-2022} present a 3D change detection method based on density adaptive local Euclidean distance, which includes calculating the local Euclidean distances from each point, improving the local geometric Euclidean distance based on the local density, and clustering the change detection results using Euclidean clustering. For the \textit{region-based methods}, CHM-CD~\cite{8272508} introduces a method that first detects the large changes by comparing the canopy height models from the two LiDAR data, and then an object-based technique is utilized to get the change detection results.

 \par In conclusion, pixel-based methods include their ease of implementation, computational efficiency, and ability to identify small changes at high resolution, making them well-suited for detecting changes in homogeneous regions. However, these methods are not suitable for detecting changes in heterogeneous regions, do not capture spatial information, and are sensitive to noise. Region-based methods are robust to noise and suitable for detecting changes in heterogeneous regions, as well as capturing spatial information such as the shape and size of changes, making them well-suited for detecting changes in regions with complex spectral properties. The disadvantages of region-based methods for change detection include their computational intensity, sensitivity to the choice of segmentation algorithm, and potential ineffectiveness in detecting small changes, which may require expert knowledge to select appropriate segmentation parameters. Hybrid methods combine the strengths of pixel-based and region-based methods, making them effective in detecting small and large changes in complex scenes, as well as robust to noise and suitable for detecting changes in heterogeneous regions. Hybrid methods offer advantages such as effectiveness in detecting small and large changes in complex scenes, robustness to noise, and suitability for detecting changes in heterogeneous regions, while they have limitations such as computational intensity, sensitivity to the choice of segmentation algorithm, and potential ineffectiveness in detecting changes in regions with complex spectral properties, which may require a high degree of expertise to implement.

\begin{table*}[!ht]
  \scriptsize
  \centering
  \caption{Representative literature on various supervision modes for different data sources in remote sensing change detection tasks. We also illustrate the fusion category for each corresponding approach.}
  \scalebox{1.0}{
  \begin{tabular}
  {p{0.15\textwidth}|p{0.1\textwidth}|p{0.11\textwidth}|p{0.05\textwidth}|p{0.47\textwidth}}
    %\toprule %{c|l|m{7cm}}
    \hline\hline
    %\hline
  \rowcolor{gray!60!}  \makecell[c]{\textbf{Method}} & \makecell[c]{\textbf{Source}} & \makecell[c]{\textbf{Category}} & \makecell[c]{\textbf{Fusion}} & \makecell[c]{\textbf{Highlights}}\\
  \hline \hline
  \rowcolor{gray!30!} \multicolumn{5}{c} {\textbf{SAR Methods}}  \\
    \hline
    \makecell[c]{SFCNet~\cite{DBLP:journals/tgrs/ZhangJLYSL22}} & \makecell[c]{TGRS 2022} & \makecell[c]{Unsupervised} & \makecell[c]{Early} & \makecell[l]{proposes a sparse feature clustering network for unsupervised change detection \\ in SAR images.} \\
    \hline
    \makecell[c]{HFEM~\cite{9829884}} & \makecell[c]{TGRS 2022} & \makecell[c]{Unsupervised} & \makecell[c]{Early}& \makecell[l]{introduces an unsupervised change detection method that contains three procedures: \\ difference image generation, thresholding, and spatial analysis.} \\
    \hline
    \makecell[c]{CycleGAN-CD~\cite{DBLP:journals/tgrs/SahaBB21}} & \makecell[c]{TGRS 2021} & \makecell[c]{Unsupervised} & \makecell[c]{Middle}& \makecell[l]{introduces a SAR change detection method based on cycle-consistent generative \\ adversarial network.} \\
    \hline
    \makecell[c]{Two-step~\cite{DBLP:journals/lgrs/WangSG22}} & \makecell[c]{GRSL 2022} & \makecell[c]{Semi-supervised} & \makecell[c]{Middle}& \makecell[l]{presents a two-step semi-supervised model based on representation learning and \\ pseudo labels.} \\
    \hline
    \makecell[c]{LCS-EnsemNet~\cite{DBLP:journals/staeors/WangWCL21}} & \makecell[c]{J-STARS 2021} & \makecell[c]{Semi-supervised} & \makecell[c]{Early}& \makecell[l]{develops a semi-supervised method with two separate branches by incorporating \\ a label-consistent self-ensemble network.} \\
    \hline
    \makecell[c]{SSN~\cite{DBLP:journals/lgrs/ChenZWY23}} & \makecell[c]{GRSL 2023} & \makecell[c]{Supervised} & \makecell[c]{Early}& \makecell[l]{proposes a Stockwell scattering network that combines wavelet scattering network \\ and Fourier scattering network.} \\
    \hline
    \makecell[c]{STGCNet~\cite{DBLP:journals/lgrs/ZhangSYW22}} & \makecell[c]{GRSL 2022} & \makecell[c]{Supervised} & \makecell[c]{Early}& \makecell[l]{introduces a deep spatial-temporal gray-level co-occurrence aware convolutional \\ neural network.} \\
    \hline\hline
  \rowcolor{gray!30!}  \multicolumn{5}{c} {\textbf{Multi-spectral Methods}}  \\
    \hline
    \makecell[c]{CAE~\cite{DBLP:journals/tgrs/BergamascoSBB22}} & \makecell[c]{TGRS 2022} & \makecell[c]{Unsupervised} & \makecell[c]{Late}& \makecell[l]{proposes an unsupervised change detection method that exploits multiresolution \\ deep feature maps derived by a convolutional autoencoder.} \\
    \hline
    \makecell[c]{PixSSLs~\cite{DBLP:journals/tgrs/ChenB22b}} & \makecell[c]{TGRS 2022} & \makecell[c]{Unsupervised} & \makecell[c]{Late}& \makecell[l]{introduces a pixel-wise contrastive approach with pseudo-Siamese network.} \\
    \hline
    \makecell[c]{GAN-CD~\cite{DBLP:journals/tgrs/RenWGZC21}} & \makecell[c]{TGRS 2021} & \makecell[c]{Unsupervised} & \makecell[c]{Late}& \makecell[l]{introduces a GAN-based procedure for unsupervised change detection in \\ satellite images.} \\
    \hline
    \makecell[c]{SSALN~\cite{DBLP:journals/tgrs/ShiWQLJ22}} & \makecell[c]{TGRS 2022} & \makecell[c]{Semi-supervised} & \makecell[c]{Late}& \makecell[l]{proposes a semi-supervised adaptive ladder network for change detection in remote \\ sensing images.} \\
    \hline
    \makecell[c]{RCL~\cite{9978928}} & \makecell[c]{TGRS 2022} & \makecell[c]{Semi-supervised} & \makecell[c]{Late}& \makecell[l]{proposes a reliable contrastive learning method for semi-supervised remote sensing \\ image change detection.} \\
    \hline
    \makecell[c]{DifUnet++~\cite{DBLP:journals/lgrs/ZhangYGYSYZ22}} & \makecell[c]{GRSL 2022} & \makecell[c]{Supervised} & \makecell[c]{Early}& \makecell[l]{proposes an effective satellite image change detection network based on Unet++ \\ and differential pyramid.} \\
    \hline
    \makecell[c]{SDMNet~\cite{DBLP:journals/lgrs/LiYZM22}} & \makecell[c]{GRSL 2022} & \makecell[c]{Supervised} & \makecell[c]{Late}& \makecell[l]{proposes a deep-supervised dual discriminative metric network that is trained \\ end-to-end for change detection in high-resolution images.} \\
    \hline\hline
  \rowcolor{gray!30!}  \multicolumn{5}{c} {\textbf{Hyperspectral Methods}}  \\
    \hline
    \makecell[c]{MD-HSI-CD~\cite{DBLP:journals/staeors/LiGDLHWFHTLM21}} & \makecell[c]{J-STARS 2021} & \makecell[c]{Unsupervised} & \makecell[c]{Early}& \makecell[l]{proposes an unsupervised end-to-end framework that employs two model-driven \\ methods for hyperspectral image change detection task.} \\
    \hline
    \makecell[c]{BCG-Net~\cite{DBLP:journals/tip/HuWDZ23}} & \makecell[c]{TIP 2022} & \makecell[c]{Unsupervised} & \makecell[c]{Middle}& \makecell[l]{introduces an unsupervised hyperspectral multiclass change detection network \\ based on binary change detection approaches.} \\
    \hline
    \makecell[c]{$\text{S}^{2}$MCD~\cite{DBLP:conf/igarss/LiuTBD17}} & \makecell[c]{IGARSS 2017} & \makecell[c]{Semi-supervised} & \makecell[c]{Early}& \makecell[l]{proposes a new semi-supervised framework that combines unsupervised change \\ representation technique and supervised classifiers.} \\
    \hline
    \makecell[c]{RSCNet~\cite{DBLP:journals/tgrs/WangWWB22}} & \makecell[c]{TGRS 2022} & \makecell[c]{Supervised} & \makecell[c]{Early}& \makecell[l]{proposes an end-to-end residual self-calibrated network to increase the accuracy \\ of hyperspectral change detection task.} \\
    \hline
    \makecell[c]{MP-ConvLSTM~\cite{DBLP:journals/tgrs/ShiZZZX22}} & \makecell[c]{TGRS 2022} & \makecell[c]{Supervised} & \makecell[c]{Late}& \makecell[l]{proposes a multipath convolutional long short-term memory and multipath \\ convolutional LSTM for hyperspectral image change detection task.} \\
    \hline\hline
  \rowcolor{gray!30!}  \multicolumn{5}{c} {\textbf{Heterogeneous Methods}}  \\
    \hline
    \makecell[c]{BASNet~\cite{DBLP:journals/lgrs/JiaZLZW22}} & \makecell[c]{GRSL 2022} & \makecell[c]{Unsupervised} & \makecell[c]{Late}& \makecell[l]{introduces a new bipartite adversarial autoencoder with structural self-similarity \\ for heterogeneous images.} \\
    \hline
    \makecell[c]{ACE-Net~\cite{DBLP:journals/tgrs/LuppinoKBMSJA22}} & \makecell[c]{TGRS 2022} & \makecell[c]{Unsupervised} & \makecell[c]{Late}& \makecell[l]{introduces two new network architectures trained with loss functions weighted \\ by priors that reduce the impact of change pixels on the learning objective.} \\
    \hline
    \makecell[c]{$\text{S}^{3}$N~\cite{DBLP:journals/tgrs/JiangLZH22}} & \makecell[c]{TGRS 2022} & \makecell[c]{Semi-supervised} & \makecell[c]{Middle}& \makecell[l]{presents a new semi-supervised Siamese network based on transfer learning.} \\
    \hline
    \makecell[c]{M-UNet~\cite{DBLP:journals/lgrs/LvHGBZS22}} & \makecell[c]{GRSL 2022} & \makecell[c]{Supervised} & \makecell[c]{Early}& \makecell[l]{introduces a heterogeneous image change detection task based on classical UNet.} \\
    \hline
    \makecell[c]{DHFF~\cite{DBLP:journals/staeors/JiangLLZH20}} & \makecell[c]{J-STARS 2020} & \makecell[c]{Supervised} & \makecell[c]{Middle}& \makecell[l]{presents a new deep homogeneous feature fusion for heterogeneous image change \\ detection based on image style transfer.} \\
    \hline\hline
      \rowcolor{gray!30!}  \multicolumn{5}{c} {\textbf{3D Change Detection Methods}}  \\
    \hline
    \makecell[c]{CamShift~\cite{DBLP:journals/staeors/XiaoXEV16}} & \makecell[c]{J-STARS 2016} & \makecell[c]{Unsupervised} & \makecell[c]{Late}& \makecell[l]{proposes a Pollock model with CamShift algorithm to segment connected \\ components into individual trees.} \\
    \hline 
    \makecell[c]{CVA-CD~\cite{DBLP:journals/lgrs/MarinelliCBB22}} & \makecell[c]{GRSL 2022} & \makecell[c]{Unsupervised} & \makecell[c]{Late}& \makecell[l]{proposes an unsupervised change detection algorithm of lidar data based on polar \\ change vector analysis.} \\
    \hline 
   \makecell[c]{Dual Stream~\cite{DBLP:journals/corr/abs-2204-12535}} & \makecell[c]{Arxiv 2022} & \makecell[c]{Supervised} & \makecell[c]{Middle}& \makecell[l]{presents a UNet model for segmenting the buildings from the \\ background.} \\
    \hline 
   \makecell[c]{Siamese KPConv~\cite{DEGELIS2023274}} & \makecell[c]{ISPRS \\ JPRS 2023} & \makecell[c]{Supervised} & \makecell[c]{Middle}& \makecell[l]{proposes a deep Siamese KPConv network that deals with raw 3D point cloud data \\ to perform change detection and categorization.} \\
    \hline\hline  
  \end{tabular}}
  \label{tab:data_supervisions}
\end{table*}

\subsection{Taxonomy based on Supervision Modes}
%\xt{I think we should summarize the common goals or ideas for each sub-aspects and highlight the motivation of these common approaches.}
%\xt{The connection of each method is a little weak.}
\par Based on the supervision modes, the existing change detection algorithms can be categorized into three types: unsupervised learning, semi-supervised learning, and supervised learning as follows. 
\begin{itemize}
\item \textbf{Unsupervised learning.} Unsupervised learning is a machine learning technique that discovers patterns and structures in data without guidance or labels, enabling the identification of hidden relationships and structures without prior knowledge. However, it can be difficult to interpret. It may suffer from the ``curse of dimensionality'', where the number of features or dimensions of the data can lead to computational inefficiencies or inaccurate results. 
\item \textbf{Semi-supervised learning.} Semi-supervised learning aims at training the algorithm with a limited amount of labeled data and a large set of unlabeled data. It is advantageous when labeled data is scarce or expensive to obtain and can help improve model accuracy by leveraging unlabeled data. Still, its implementation can be difficult, and its performance depends on the quality of unlabeled data, which can introduce noise and lead to decreased performance. \item \textbf{Supervised learning.} Supervised learning is trained using labeled data to make accurate predictions on new, unseen data by learning patterns from input and corresponding output data. It is easy to implement with readily available labeled data and can be used to solve various problems, but it requires a large amount of labeled data that should be accurate and unbiased, and models may struggle with data different from the training data, leading to overfitting or underfitting. 
\end{itemize}
\par In the following, we will provide detailed and practical insights into change detection algorithms for each data source, based on the modes of supervision.

\par \noindent \textbf{1) SAR Methods.} Due to the scarcity of publicly available SAR image change detection datasets, \textit{unsupervised learning methods} have become prevalent. In particular, image clustering techniques and parameter-fixed feature extraction networks are commonly utilized. For instance, SFCNet~\cite{DBLP:journals/tgrs/ZhangJLYSL22} proposes a sparse feature clustering network for detecting changes in SAR images, which is pre-trained with the multi-objective sparse feature learning model. HFEM~\cite{9829884} introduces a new change detection method for very few changes or even none changed areas, which contains difference image generation, a thresholding method, and one conditional random fields method. CycleGAN-CD~\cite{DBLP:journals/tgrs/SahaBB21} presents a SAR change detection method based on cycle-consistent generative adversarial network. For the \textit{semi-supervised learning approaches}, Wang \textit{et al.}~\cite{DBLP:journals/lgrs/WangSG22} propose a patch-based semi-supervised method to detect changed pixels from limited training data, including the unsupervised pretraining and iterative discrimination. LCS-EnsemNet~\cite{DBLP:journals/staeors/WangWCL21} develops a semi-supervised method based on a two-branch strategy by incorporating a label-consistent self-ensemble network. Within the category of \textit{supervised learning approaches}, SSN~\cite{DBLP:journals/lgrs/ChenZWY23} proposes a Stockwell scattering network that combines a wavelet scattering network and a Fourier scattering network. Zhang~\textit{et al.}~\cite{DBLP:journals/lgrs/ZhangSYW22} introduce a deep spatial-temporal gray-level co-occurrence aware CNNs, which can effectively mine the spatial-temporal information and obtain robust results.
 
\par \noindent \textbf{2) Multi-spectral Methods.} For the \textit{unsupervised learning-based approaches}, Bergamasco \textit{et al.}~\cite{DBLP:journals/tgrs/BergamascoSBB22} introduce a new approach for unsupervised change detection, which leverages multi-resolution deep feature maps obtained from a convolutional autoencoder. Chen \textit{et al.}~\cite{DBLP:journals/tgrs/ChenB22b} propose a pseudo-siamese network that is trained to obtain pixel-wise representations and to align features from shifted image pairs. Ren \textit{et al.}~\cite{DBLP:journals/tgrs/RenWGZC21} propose a new unsupervised change detection framework utilizing a generative adversarial network to generate many better-coregistered images. For the \textit{semi-supervised learning algorithms}, SSALN~\cite{DBLP:journals/tgrs/ShiWQLJ22} proposes a semi-supervised adaptive ladder network for change detection in remote sensing images, which can update pseudo labels iteratively. Wang \textit{et al.}~\cite{9978928} propose a reliable contrastive learning method for semi-supervised remote sensing image change detection by selecting reliable samples according to the prediction uncertainty of unlabeled images and introducing the contrastive loss. Among the set of \textit{supervised learning models}, Zhang \textit{et al.}~\cite{DBLP:journals/lgrs/ZhangYGYSYZ22} proposes an effective satellite images change detection network DifUnet++, which takes a differential pyramid of two input images as the input and incorporates a side-out fusion strategy to predict the detection results. SDMNet~\cite{DBLP:journals/lgrs/LiYZM22} introduces a new end-to-end metric learning algorithm for remote sensing change detection tasks, which introduces a discriminative decoder network to aggregate multiscale and global contextual information to obtain discriminative consistent features and a discriminative implicit metric module to measure the distance between features to achieve the changes.

\par \noindent \textbf{3) Hyperspectral Methods.} Amongst \textit{unsupervised learning techniques}, Li~\textit{et al.}~\cite{DBLP:journals/staeors/LiGDLHWFHTLM21} propose an unsupervised end-to-end framework that employs two model-driven methods for hyperspectral image change detection task. BCG-Net~\cite{DBLP:journals/tip/HuWDZ23} proposes an unsupervised hyperspectral multiclass change detection network based on the binary change detection approach, which aims to boost the multiclass change detection result and unmixing result. For the \textit{semi-supervised learning algorithms}, Liu \textit{et al.}~\cite{DBLP:conf/igarss/LiuTBD17} introduce the semi-supervised hyperspectral detection approach, which generates the pseudo label samples from the state-of-the-art unsupervised change representation technique, and classifies the changed regions and no-changed regions by a supervised classifier. Within the category of \textit{supervised learning algorithms}, Wang \textit{et al.}~\cite{DBLP:journals/tgrs/WangWWB22} introduce a residual self-calibrated network for hyperspectral image change detection task, which adaptively builds inter-spatial and inter-spectral dependencies around each spatial location with fewer extra parameters and reduced complexity. Shi \textit{et al.}~\cite{DBLP:journals/tgrs/ShiZZZX22} propose a multipath convolutional long short-term memory neural network for hyperspectral change detection task, which introduces an efficient channel attention module to refine features of different paths.
 
\par \noindent \textbf{4) Heterogeneous Methods.} For the \textit{unsupervised learning methods}, BASNet~\cite{DBLP:journals/lgrs/JiaZLZW22} proposes a new bipartite adversarial autoencoder with structural self-similarity for heterogeneous remote sensing images, which introduces a structural consistency loss to transform the images into a common domain, and an adversarial loss to make image translation with a more consistent style. Luppino \textit{et al.}~\cite{DBLP:journals/tgrs/LuppinoKBMSJA22} propose two novel network architectures that are trained using loss functions weighted by priors, which helps to minimize the effect of changing pixels on the overall learning objective. Within the category of \textit{semi-supervised learning approaches}, $\text{S}^{3}$N~\cite{DBLP:journals/tgrs/JiangLZH22} proposes a new semi-supervised siamese network based on transfer learning, which takes the low- and high-level features separately and treats them differently. Amongst supervised learning techniques, Lv \textit{et al.}~\cite{DBLP:journals/lgrs/LvHGBZS22} introduces a UNet model for the heterogeneous remote sensing image change detection task, which incorporates a multiscale convolution module into a U-Net backbone to cover the various sizes and shapes of ground targets. DHFF~\cite{DBLP:journals/staeors/JiangLLZH20} proposes a deep homogeneous feature fusion method for heterogeneous image change detection tasks, which segregates the semantic content and the style features to perform the homogeneous transformation.

\par \noindent \textbf{5) 3D Change Detection Methods.} Within the \textit{unsupervised learning methods}, Xiao \textit{et al.}~\cite{DBLP:journals/staeors/XiaoXEV16} propose a tree-shaped model to continuously adaptive mean shift algorithm to classify the clustered components into individual trees, then the tree parameters are derived with a point-based method and a model-based method. Marinelli \textit{et al.}~\cite{DBLP:journals/lgrs/MarinelliCBB22} introduce a new unsupervised change detection algorithm in lidar point clouds, which utilizes a polar change vector analysis to automatically discriminate between the different classes of change. Among the \textit{supervised learning algorithms}, Yadav \textit{et al.}~\cite{DBLP:journals/corr/abs-2204-12535} propose a change detection model with U-Net for segmenting the buildings from the background, which utilizes an automatic method to reduce the 3D point clouds into a much smaller representation without losing necessary information. Siamese KPConv~\cite{DEGELIS2023274} presents a Siamese network to perform 3D point cloud change detection and categorization in a single step.

\par In summary, unsupervised learning methods excel in detecting changes without labeled data and accommodating diverse data sources, but their incapability of distinguishing true changes from noise restricts their performance in complex environments. Semi-supervised learning methods leverage both labeled and unlabeled data to achieve better accuracy than unsupervised methods, but their effectiveness relies on the quality and quantity of labeled data. Supervised learning methods can achieve high accuracy in detecting changes with sufficient labeled data, but their inflexibility and dependence on labeled data can pose limitations in some scenarios, such as complex patterns and scarce labeled data.
  \begin{table*}[!ht]
  \scriptsize
  \centering
  \caption{Representative literature on various learning frameworks for different data sources in remote sensing change detection tasks. We also illustrate the fusion category for each corresponding approach.}
  \scalebox{1.0}{
  \begin{tabular}
  {p{0.13\textwidth}|p{0.1\textwidth}|p{0.08\textwidth}|p{0.05\textwidth}|p{0.50\textwidth}}
    %\toprule %{c|l|m{7cm}}
    \hline\hline
    %\hline
  \rowcolor{gray!60!}  \makecell[c]{\textbf{Method}} & \makecell[c]{\textbf{Source}} & \makecell[c]{\textbf{Category}} & \makecell[c]{\textbf{Fusion}}& \makecell[c]{\textbf{Highlights}}\\
  \hline \hline
  \rowcolor{gray!30!}  \multicolumn{5}{c} {\textbf{SAR Methods}}  \\
    \hline
    \makecell[c]{Yu \textit{et al.}~\cite{DBLP:journals/lgrs/YuY22}} & \makecell[c]{GRSL 2022} & \makecell[c]{Traditional}  & \makecell[c]{Early} & \makecell[l]{proposes a traditional change detection approach that combines a symmetric similarity \\ matrix, a Shannon entropy, and an image segmentation method based on MRF. } \\
    \hline
    \makecell[c]{Liu \textit{et al.}~\cite{DBLP:journals/lgrs/LiuCL22}} & \makecell[c]{GRSL 2022} & \makecell[c]{Traditional} & \makecell[c]{Middle}& \makecell[l]{proposes an unsupervised method to automatically select training samples and utilizes a \\ well-trained RF classifier to achieve change detection result.  } \\
    \hline
    \makecell[c]{Vinholi \textit{et al.} ~\cite{DBLP:journals/tgrs/VinholiPSMP22}} & \makecell[c]{TGRS 2022} & \makecell[c]{CNN} & \makecell[c]{Early}& \makecell[l]{presents two supervised change detection algorithms based on CNNs that use stacks \\ of SAR images.} \\
    \hline
    \makecell[c]{DDNet~\cite{DBLP:journals/lgrs/QuGDDL22}} & \makecell[c]{GRSL 2022} & \makecell[c]{CNN} & \makecell[c]{Middle}& \makecell[l]{presents a dual-domain network to jointly exploit the spatial and frequency features for \\ SAR change detection task.} \\
    \hline
    \makecell[c]{MSDC~\cite{DBLP:journals/tgrs/DongMJLL22}} & \makecell[c]{TGRS 2022} & \makecell[c]{A or T} & \makecell[c]{Middle}& \makecell[l]{proposes a unified framework that integrates unsupervised clustering with CNN to learn \\ clustering-friendly feature representations} \\
    \hline
    \makecell[c]{MACNet~\cite{DBLP:conf/igarss/LiGDQ22}} & \makecell[c]{IGARSS 2022} & \makecell[c]{A or T} & \makecell[c]{Early}& \makecell[l]{introduces a multi-scale attention convolution network to exploit the spatial information of \\ feature maps from different scales.} \\ 
    \hline
    \makecell[c]{ASGF~\cite{DBLP:journals/lgrs/ZhaoMWOMW23}} & \makecell[c]{GRSL 2023} & \makecell[c]{A or T} & \makecell[c]{Early}& \makecell[l]{proposes a new SAR image change detection algorithm that is based on an attention \\ mechanism in the spatial domain and a gated linear unit in the frequency domain.} \\
    \hline\hline
  \rowcolor{gray!30!}  \multicolumn{5}{c} {\textbf{Multi-spectral Methods}}  \\
    \hline
    \makecell[c]{MDF CD~\cite{DBLP:journals/lgrs/ShaoYLDR22}} & \makecell[c]{GRSL 2022} & \makecell[c]{Traditional} & \makecell[c]{Early}& \makecell[l]{presents a novel multiscale decision fusion method for unsupervised  change detection \\ approach based on Dempster–Shafer theory and modified conditional random field.} \\
    \hline
    \makecell[c]{Fang \textit{et al.}~\cite{DBLP:journals/lgrs/FangDWLT22}} & \makecell[c]{GRSL 2022} & \makecell[c]{Traditional} & \makecell[c]{Early}& \makecell[l]{proposes an unsupervised change detection method for high spatial resolution images based \\ on the weighted change vector analysis and the improved Markov random field.} \\
    \hline
    \makecell[c]{ECFNet~\cite{10023508}} & \makecell[c]{GRSL 2023} & \makecell[c]{CNN} & \makecell[c]{Middle}& \makecell[l]{presents a simple and efficient network architecture, extraction, comparison, and fusion \\ network for change detection in remote-sensing images.} \\
    \hline
    \makecell[c]{Chen \textit{et al.}~\cite{DBLP:journals/tgrs/ChenLCS22}} & \makecell[c]{TGRS 2022} & \makecell[c]{CNN} & \makecell[c]{Late}& \makecell[l]{incorporates semantic supervision into the self-supervised learning framework for remote \\ sensing image change detection.} \\
    \hline
    \makecell[c]{DMATNet~\cite{DBLP:journals/tgrs/SongHL22}} & \makecell[c]{TGRS 2022} & \makecell[c]{A or T} & \makecell[c]{Middle}& \makecell[l]{presents a dual-feature mixed attention-based
transformer network for remote \\ sensing image change detection.} \\
    \hline
    \makecell[c]{Chen \textit{et al.}~\cite{DBLP:journals/tgrs/ChenQS22}} & \makecell[c]{TGRS 2022} & \makecell[c]{A or T} & \makecell[c]{Middle}& \makecell[l]{proposes a bitemporal image transformer to efficiently and effectively model contexts \\ within the spatial-temporal domain.} \\
    \hline 
    \makecell[c]{PA-Former~\cite{DBLP:journals/lgrs/LiuSCL22}} & \makecell[c]{GRSL 2022} & \makecell[c]{A or T} & \makecell[c]{Middle}& \makecell[l]{introduces an end-to-end PA-Former for building change detection that combines prior \\ extraction and contextual fusion together.} \\
    \hline
    \makecell[c]{ACAHNet~\cite{DBLP:journals/tgrs/ZhangCWL23}} & \makecell[c]{TGRS 2023} & \makecell[c]{A or T} & \makecell[c]{Middle}& \makecell[l]{proposes an asymmetric cross-attention hierarchical network by combining CNN and \\ transformer in a series-parallel manner.} \\
    \hline\hline 
  \rowcolor{gray!30!}  \multicolumn{5}{c} {\textbf{Hyperspectral Methods}}  \\
    \hline 
    \makecell[c]{ACDA~\cite{DBLP:journals/staeors/HuWZD21}} & \makecell[c]{J-STARS 2021} & \makecell[c]{Traditional} & \makecell[c]{Late}& \makecell[l]{proposes a hyperspectral anomaly change detection algorithm based on auto-encoder to \\ enhance nonlinear representation.} \\
    \hline
    \makecell[c]{SMSL~\cite{DBLP:journals/tgrs/ChangKG22}} & \makecell[c]{TGRS 2022} & \makecell[c]{Traditional} & \makecell[c]{Middle}& \makecell[l]{introduces a sketched multiview subspace learning model for hyperspectral image \\ anomalous change detection task.} \\
    \hline
    \makecell[c]{MMSRC~\cite{DBLP:journals/staeors/GeTBZXS22}} & \makecell[c]{J-STARS 2022} & \makecell[c]{CNN} & \makecell[c]{Early}& \makecell[l]{proposes a new multidirection and multi-scale spectral-spatial residual network \\ for hyperspectral multiclass change detection.} \\
    \hline 
    \makecell[c]{SFBS-FFGNET \\ ~\cite{DBLP:journals/tgrs/OuLTZX22}} & \makecell[c]{TGRS 2022} & \makecell[c]{CNN} & \makecell[c]{Early}& \makecell[l]{proposes a CNN framework involving slow-fast band selection and feature fusion \\ grouping for hyperspectral image change detection.} \\
    \hline
    \makecell[c]{SST-Former~\cite{DBLP:journals/tgrs/WangHSGLZR22}} & \makecell[c]{TGRS 2022} & \makecell[c]{A or T} & \makecell[c]{Middle}& \makecell[l]{proposes a joint spectral, spatial, and temporal transformer for hyperspectral image \\ change detection.} \\
    \hline
    \makecell[c]{Dong \textit{et al.}~\cite{DBLP:journals/tgrs/DongZQXLHL23}} & \makecell[c]{TGRS 2023} & \makecell[c]{A or T} & \makecell[c]{Middle}& \makecell[l]{proposes an abundance matrix correlation analysis network based on hierarchical \\ multihead self-cross hybrid attention for hyperspectral change detection.} \\
    \hline
    \makecell[c]{CSANet~\cite{DBLP:journals/lgrs/SongNCW22}} & \makecell[c]{GRSL 2022} & \makecell[c]{A or T} & \makecell[c]{Middle}& \makecell[l]{proposes a new cross-temporal interaction symmetric attention network.} \\
    \hline
    \makecell[c]{DPM$\text{s}^{2}$raN~\cite{DBLP:journals/tgrs/YangQXDLD22}} & \makecell[c]{TGRS 2022} & \makecell[c]{A or T} & \makecell[c]{Middle}& \makecell[l]{proposes a deep multiscale pyramid network with spatial–spectral residual attention.} \\
    \hline\hline
  \rowcolor{gray!30!}  \multicolumn{5}{c} {\textbf{Heterogeneous Methods}}  \\
    \hline
    \makecell[c]{SDA-HCD~\cite{DBLP:journals/tgrs/SunLGKL22}} & \makecell[c]{TGRS 2022} & \makecell[c]{Traditional} & \makecell[c]{Late}& \makecell[l]{introduces a spectral domain analysis for heterogeneous change detection.} \\
    \hline
    \makecell[c]{Sun \textit{et al.}~\cite{DBLP:journals/tgrs/SunLGLK22}} & \makecell[c]{TGRS 2022} & \makecell[c]{Traditional} & \makecell[c]{Middle}& \makecell[l]{proposes an unsupervised image regression method for change detection tasks based on \\ the structure consistency.} \\
    \hline
    \makecell[c]{CAE~\cite{DBLP:journals/tnn/WuLYQMG22}} & \makecell[c]{TNNLS 2022} & \makecell[c]{CNN} & \makecell[c]{Late}& \makecell[l]{proposes an unsupervised change detection method that contains a convolutional \\ autoencoder and a commonality autoencoder.} \\
    \hline 
    \makecell[c]{TVRBN~\cite{DBLP:journals/tgrs/HuLX22}} & \makecell[c]{TGRS 2022} & \makecell[c]{CNN} & \makecell[c]{Middle}& \makecell[l]{proposes an unsupervised joint learning model based on a total variation regularization \\ and bipartite CNNs.} \\
    \hline
    \makecell[c]{DA-MSCDNet ~\cite{DBLP:journals/aeog/ZhangFHTPLCY22}} & \makecell[c]{IJAEOG 2022} & \makecell[c]{CNN} & \makecell[c]{Middle}& \makecell[l]{introduces a domain adaptation and a multi-source change detection network to process \\ heterogeneous images.} \\
    \hline
    \makecell[c]{TSCNet~\cite{DBLP:journals/remotesensing/WangCFS23}} & \makecell[c]{Remote Sensing \\ 2023} & \makecell[c]{A or T} & \makecell[c]{Middle}& \makecell[l]{proposes a new topology-coupling algorithm for heterogeneous image change \\ detection task.} \\
    \hline\hline
      \rowcolor{gray!30!}  \multicolumn{5}{c} {\textbf{3D Change Detection Methods}}  \\
    \hline
    \makecell[c]{Dai \textit{et al.}~\cite{DBLP:journals/remotesensing/DaiZL20}} & \makecell[c]{Remote Sensing \\ 2020} & \makecell[c]{Traditional} & \makecell[c]{Middle}& \makecell[l]{presents an unsupervised, object-based method for integrated building extraction and \\ change detection with point cloud data.} \\
    \hline
    \makecell[c]{Liu \textit{et al.}~\cite{DBLP:journals/ijgi/LiuLWW21}} & \makecell[c]{ISPRS IJGI 2021} & \makecell[c]{Traditional} & \makecell[c]{Early}& \makecell[l]{introduces an approach for 3D change detection using point-based comparison.} \\
   % \hline 
   %\makecell[c]{Siamese KPConv \\ ~\cite{DEGELIS2023274}} & \makecell[c]{ISPRS \\ JPRS 2023} & \makecell[c]{CNN} & \makecell[c]{Early}& \makecell[l]{proposes a deep Siamese KPConv network that deals with raw 3D point cloud data \\ to perform change detection and categorization.} \\
    \hline
    \makecell[c]{ChangeGAN~\cite{DBLP:journals/ral/NagyKB21}} & \makecell[c]{RAL 2021} & \makecell[c]{CNN} & \makecell[c]{Middle}& \makecell[l]{proposes a generative adversarial network architecture for point cloud change detection.} \\
    \hline\hline
  \end{tabular}}
  \label{tab:data_learning_strategy}
\end{table*}

  \subsection{Taxonomy based on Learning Frameworks}
  \par This subsection presents a taxonomy of the existing change detection algorithms based on their learning frameworks. The taxonomy is divided into three categories: traditional methods, CNN-based methods, and attention or transformer-based methods. Traditional methods commonly utilize conventional clustering or classification algorithms, such as SVM~\cite{doi:10.1080/01431161.2014.951740, DBLP:journals/staeors/HuoCDZP16, DBLP:journals/tgrs/BovoloBM08}, random forests~\cite{DBLP:journals/ijgi/SeoKELP18}, and MRF~\cite{DBLP:journals/tgrs/BruzzoneF00, DBLP:journals/tip/TouatiMD20}, to partition the data into changed and unchanged areas. In contrast, CNN-based methods use CNNs to automatically learn feature representations from the data, enabling the capture of complex patterns and spatial dependencies between pixels. This approach has demonstrated improved performance over traditional methods. Attention or transformer-based methods, also based on deep learning methods, employ self-attention mechanisms to weight different regions of the input images based on their relevance to the task, or to capture long-range dependencies between pixels and generate feature representations. These methods have demonstrated superior performance compared to both CNN-based methods and traditional methods, representing the new state-of-the-art for change detection in remote sensing imagery.
  %\par In this subsection, we will classify the existing change detection algorithms according to the learning frameworks into the following categories: traditional methods, CNN-based methods, attention or transformer-based methods. Traditional methods usually employ the existing conventional clustering or classification algorithms, as well as some statistical analysis strategies to partition the data into changed areas and unchanged areas. CNN-based methods use convolutional neural networks to automatically learn feature representations from the data. They have shown improved performance over traditional methods, as they can capture complex patterns and spatial dependencies between pixels. Attention or transformer-based methods, which are also based on deep learning methods, utilize the self-attention mechanisms to weight different regions of the input images based on their relevance to the task or capture long-range dependencies between pixels and generate feature representations. These methods have shown new state-of-the-art performance than the CNN-based methods and traditional methods.
 \par \noindent \textbf{1) SAR Methods.} Among the \textit{traditional methods}, Yu \textit{et al.}~\cite{DBLP:journals/lgrs/YuY22} propose an unsupervised change detection algorithm, which combines a symmetric similarity matrix based on a likelihood ratio test, a Shannon entropy to calculate the difference image, and an image segmentation method based on MRF. Liu \textit{et al.}~\cite{DBLP:journals/lgrs/LiuCL22} introduce an existing detection result to select the training data, a random forest classifier to achieve classification results, and median filtering to eliminate singular points. Within the \textit{CNN-based approaches}, Vinholi \textit{et al.}~\cite{DBLP:journals/tgrs/VinholiPSMP22} present two supervised change detection algorithms based on CNNs, which consist of the following four stages: difference image formation, semantic segmentation, clustering, and change classification. DDNet~\cite{DBLP:journals/lgrs/QuGDDL22} proposes a new SAR change detection method that uses features from both spatial and frequency domains. Specifically, a multi-region convolution module is utilized to enhance the spatial features, and a discrete cosine transform and gating mechanism is employed to extract frequency features.  In the category of \textit{attention or transformer-based algorithms}, MSDC~\cite{DBLP:journals/tgrs/DongMJLL22} proposes an unsupervised change detection framework by combining K-means++ clustering and deep convolutional model, which can be jointly optimized without supervision. Li \textit{et al.}~\cite{DBLP:conf/igarss/LiGDQ22} present a multi-scale attention convolution network, which extracts the spatial information of feature maps with a linear attention weight module, and designs a linear attention weight module to emphasize the important channels adaptively, and fuses the contextual information from different scales. ASGF~\cite{DBLP:journals/lgrs/ZhaoMWOMW23} proposes an unsupervised change detection approach, which employs a clustering technique to generate pseudo labels and utilizes a convolutional neural network to enable feature learning of the network.
 
\par \noindent \textbf{2) Multi-spectral Methods.} For \textit{traditional methods}, Shao \textit{et al.}~\cite{DBLP:journals/lgrs/ShaoYLDR22} present a new multiscale decision fusion method for an unsupervised change detection approach based on Dempster–Shafer theory and modified conditional random field, which consists of difference image generation, a fuzzy clustering algorithm, fusion strategy based on Dempster–Shafer theory, and a modified CRF. Fang \textit{et al.}~\cite{DBLP:journals/lgrs/FangDWLT22} present an unsupervised approach for detecting changes in high spatial resolution images, which leverages the weighted change vector analysis technique and incorporates an improved Markov random field model. Among the set of \textit{CNN algorithms}, ECFNet~\cite{10023508} proposes a simple but efficient network for remote sensing images, which consists of a feature extraction module, a feature comparison module, and a feature fusion module. Chen \textit{et al.}~\cite{DBLP:journals/tgrs/ChenLCS22} explore the use of semantic information in a representation learning framework and propose semantic-aware pre-training based on class-balanced sampling for remote sensing image change detection. Within the category of \textit{attention or transformer techniques}, DMATNet~\cite{DBLP:journals/tgrs/SongHL22} introduces a DFMA-based transformer change detection model for high-resolution remote sensing images, which utilizes CNNs to extract coarse and fine features and employs a dual-feature mixed attention (DFMA) module to fuse these features. Chen~\textit{et al.}~\cite{DBLP:journals/tgrs/ChenQS22} propose a bitemporal image transformer to efficiently and effectively model contexts within the spatial-temporal domain for remote sensing image change detection. PA-Former~\cite{DBLP:journals/lgrs/LiuSCL22} presents a new network based on a transformer by learning a prior-aware transformer to help capture cross-temporal and long-range contextual information. ACAHNet~\cite{DBLP:journals/tgrs/ZhangCWL23} proposes an asymmetric cross-attention hierarchical network by combining CNN and transformer in a series-parallel manner, which reduces the computational complexity and enhances the interaction between features extracted from CNN and the transformer.

 \par \noindent \textbf{3) Hyperspectral Methods.} Among the \textit{traditional techniques}, ACDA~\cite{DBLP:journals/staeors/HuWZD21} introduces a new hyperspectral anomaly change detection algorithm based on a nonlinear auto-encoder, which gains better detection performance against other state-of-the-art approaches. SMSL~\cite{DBLP:journals/tgrs/ChangKG22} introduces a sketched multiview subspace learning algorithm for anomalous change detection, which preserves major information from the image pairs and improves the computational complexity using a sketched representation matrix. Within the category of \textit{CNN methods}, MMSRC~\cite{DBLP:journals/staeors/GeTBZXS22} proposes a new multi-direction and multi-scale spectral-spatial residual network for hyperspectral multiclass change detection, which improves feature variation and accuracy of hyperspectral images. SFBS-FFGNET ~\cite{DBLP:journals/tgrs/OuLTZX22} introduces a CNN framework involving slow-fast band selection and feature fusion grouping for hyperspectral image change detection, which incorporates selecting effective bands and fusing different features. For the \textit{attention or transformer algorithms}, SST-Former~\cite{DBLP:journals/tgrs/WangHSGLZR22} introduces an end-to-end transformer model for hyperspectral change detection task, which simultaneously considers the spatial, spectral, and temporal information for hyperspectral images. Dong \textit{et al.}~\cite{DBLP:journals/tgrs/DongZQXLHL23} propose an abundance matrix correlation analysis network based on hierarchical multi-head self-cross hybrid attention for hyperspectral change detection, which hierarchically highlights the correlation difference information at the subpixel level. CSANet~\cite{DBLP:journals/lgrs/SongNCW22} proposes a new cross-temporal interaction symmetric attention network, which can effectively extract and integrate the joint spatial-spectral–temporal features of the hyperspectral images, and enhance the feature discrimination ability of the changes. DPM$\text{s}^{2}$raN~\cite{DBLP:journals/tgrs/YangQXDLD22} proposes a deep multiscale pyramid network with spatial-spectral residual attention, which has a strong capability to mine multilevel and multiscale spatial–spectral features, thus improving the performance in complex changed regions.
 
\par \noindent \textbf{4) Heterogeneous Methods.} For the \textit{traditional approaches}, SDA-HCD~\cite{DBLP:journals/tgrs/SunLGKL22} introduces a spectral domain analysis-based heterogeneous change detection, which decomposes the source signal into the regressed signal and the changed signal and constrains the spectral property of the regressed signal. Sun \textit{et al.}~\cite{DBLP:journals/tgrs/SunLGLK22} propose an unsupervised image regression-based change detection method based on the structure consistency, which uses a similarity graph to translate an image, computes the difference image and then segments it into changed and unchanged classes using a superpixel-based Markovian segmentation model. Among the \textit{CNN methods}, CAE~\cite{DBLP:journals/tnn/WuLYQMG22}  proposes an unsupervised change detection method that contains only a convolutional autoencoder for feature extraction and the commonality autoencoder for commonalities exploration. TVRBN~\cite{DBLP:journals/tgrs/HuLX22} introduces an unsupervised joint learning model based on total variation regularization and bipartite CNN. DA-MSCDNet~\cite{DBLP:journals/aeog/ZhangFHTPLCY22} proposes a domain adaptation-based multi-source change detection network to process heterogeneous optical and SAR remote sensing images, which employs feature-level transformation to align inconsistent deep feature spaces. Within the category of \textit{attention or transformer algorithms}, TSCNet~\cite{DBLP:journals/remotesensing/WangCFS23} introduces a new topology-coupling-based heterogeneous remote sensing image change detection network, which transforms the feature space of heterogeneous images using an encoder-decoder structure and introduces wavelet transform, channel, and spatial attention mechanisms.

\par \noindent \textbf{5) 3D Change Detection Methods.} For the \textit{traditional algorithms}, Dai \textit{et al.}~\cite{DBLP:journals/remotesensing/DaiZL20} presents an unsupervised, object-based method for integrated building extraction and change detection using point cloud data, which combines bottom-up segmentation and clustering, as well as an object-based bidirectional algorithm. Liu \textit{et al.}~\cite{DBLP:journals/ijgi/LiuLWW21} propose an approach for 3D change detection using point-based comparison. To avoid density variation in point clouds, adaptive thresholds are calculated through the k-neighboring average distance and the local point cloud density. Among the \textit{CNN methods}, ChangeGAN~\cite{DBLP:journals/ral/NagyKB21} introduces a generative adversarial network architecture for point cloud change detection task, which combines siamese-style feature extraction, U-net-like multiscale feature extraction, and spatial transformation network blocks for optimal transformation estimation. Siamese KPConv~\cite{DEGELIS2023274} proposes a deep Siamese KPConv network that deals with raw 3D PCs to perform change detection and categorization in a single step. % \textit{Attention or transformer methods}

% Unsupervised  DBLP:journals/staeors/XiaoXEV16, DBLP:journals/tgrs/MarinelliPB18, DBLP:conf/igarss/MarinelliCBB18, doi:10.1080/17538947.2019.1585975, 8272508, isprs-annals-IV-2-W7-9-2019, isprs-annals-V-2-2020-687-2020, isprs-annals-IV-2-W5-357-2019
% Supervised MARSOCCI2023325, DEGELIS2023274,  

% Traditional DBLP:journals/staeors/XiaoXEV16, DBLP:journals/tgrs/MarinelliPB18, doi:10.1080/17538947.2019.1585975, 8272508, isprs-annals-IV-2-W7-9-2019, isprs-annals-V-2-2020-687-2020, isprs-annals-IV-2-W5-357-2019
% CNN  DEGELIS2023274, 
% Transformer MARSOCCI2023325, 
%  \end{itemize}
%  \subsection{Taxonomy based on Supervision Modes}
%  \subsubsection{Unsupervised based}
%  \subsubsection{Semi-supervised based}
%  \subsubsection{Supervised based}

 \section{Benchmark Performance} % 2-3 pages
  \label{benchmark}
  %\gl{add the benchmark performance by Yunmeng}
%\subsubsection{SAR based Benchmarks}  
\par In this section, we will provide state-of-the-art methods for change detection tasks on the dominant datasets depicted in the preliminary knowledge section. It should be noted that we only present some representative algorithms that are commonly utilized for comparative analysis.
\par \noindent \textbf{1) SAR Benchmarks.} In Tab.~\ref{tab:benchmark_SAR}, we provide a summary of several noteworthy methods on two dominant SAR datasets, i.e. the Yellow River dataset~\cite{9982693} and the Bern dataset~\cite{9829884}. Specifically, In the Yellow River dataset, the DDNet algorithm~\cite{DBLP:journals/lgrs/QuGDDL22} outperforms other methods in terms of FP and KC metrics, while the SFCNet model~\cite{DBLP:journals/tgrs/ZhangJLYSL22} achieves the best performance on FN and OA metrics. In the Bern dataset, ShearNet~\cite{9982693} and ESMOFCM~\cite{9134375} exhibit superior performance, achieving an OA metric of 99.68\%, and BIFLICM/D~\cite{DBLP:journals/lgrs/FangX22} achieves the best performance in KC metric. 
\begin{table*}[!ht]
  \scriptsize
  \centering
  \caption{Some representative methods on SAR datasets. The model with the best performance is denoted in bold.}
  \scalebox{1.0}{
  \begin{tabular}
  {p{0.15\textwidth}|p{0.10\textwidth} |p{0.06\textwidth} |p{0.06\textwidth}|p{0.06\textwidth}
  |p{0.06\textwidth}}
    %\toprule %{c|l|m{7cm}}
    \hline\hline
  \rowcolor{gray!60!}  \multicolumn{6}{c} {\textbf{SAR Methods}}  \\
    \hline\hline
  \rowcolor{gray!30!}  \makecell[c]{\textbf{Method}} & \makecell[c]{\textbf{Source}} & \makecell[c]{\textbf{FP}} & \makecell[c]{\textbf{FN}} & \makecell[c]{\textbf{OA}} & \makecell[c]{\textbf{KC}}\\  
    \hline\hline
  \rowcolor{gray!15!}   \multicolumn{6}{c}{\textbf{Yellow River Dataset}} \\
    \hline\hline
    %\makecell[c]{PCAKM~\cite{5196726}} & \makecell[c]{GRSL 2009} & \makecell[c]{955}  & \makecell[c]{1515} & \makecell[c]{97.57} & \makecell[c]{90.73} \\
    %\hline
    \makecell[c]{RFLICM~\cite{6035777}} & \makecell[c]{TIP 2012} & \makecell[c]{862} & \makecell[c]{1300} & \makecell[c]{98.33} & \makecell[c]{74.97}  \\
    \hline
    \makecell[c]{GaborPCANet~\cite{DBLP:journals/lgrs/GaoDLX16}} & \makecell[c]{GRSL 2016} & \makecell[c]{1043} & \makecell[c]{1009} & \makecell[c]{96.87} & \makecell[c]{81.21}  \\
    \hline
    \makecell[c]{CWNNs~\cite{8641484}} & \makecell[c]{GRSL 2019} & \makecell[c]{837} & \makecell[c]{1690} & \makecell[c]{96.60} & \makecell[c]{88.23}  \\
    \hline
    \makecell[c]{DCNet~\cite{DBLP:journals/staeors/GaoGDW19}} & \makecell[c]{J-STARS 2019} & \makecell[c]{790} & \makecell[c]{2137} & \makecell[c]{96.06} & \makecell[c]{86.16}  \\
    \hline
    \makecell[c]{MSAPNet~\cite{DBLP:conf/igarss/WangDCLZJ20}} & \makecell[c]{IGARSS 2020} & \makecell[c]{817} & \makecell[c]{2157} & \makecell[c]{96.00} & \makecell[c]{85.94}  \\
    \hline
    \makecell[c]{SFCNet~\cite{DBLP:journals/tgrs/ZhangJLYSL22}} & \makecell[c]{TGRS 2022} & \makecell[c]{720} & \makecell[c]{\textbf{704}} & \makecell[c]{\textbf{98.40}} & \makecell[c]{85.62}  \\
    \hline
    \makecell[c]{SSN~\cite{10016644}} & \makecell[c]{GRSL 2023} & \makecell[c]{1292} & \makecell[c]{793} & \makecell[c]{97.19} & \makecell[c]{90.66}  \\
    \hline
    \makecell[c]{DDNet~\cite{DBLP:journals/lgrs/QuGDDL22}} & \makecell[c]{GRSL 2022} & \makecell[c]{\textbf{641}} & \makecell[c]{1027} & \makecell[c]{98.36} & \makecell[c]{\textbf{93.77}}  \\
    \hline\hline

  \rowcolor{gray!15!}   \multicolumn{6}{c}{\textbf{Bern Dataset}} \\
    \hline\hline
    %\makecell[c]{PCAKM~\cite{5196726}} & \makecell[c]{GRSL 2009} & \makecell[c]{244} & \makecell[c]{\textbf{117}} & \makecell[c]{99.60} & \makecell[c]{84.78}\\
    %\hline
    \makecell[c]{GaborPCANet~\cite{DBLP:journals/lgrs/GaoDLX16}} & \makecell[c]{GRSL 2016} & \makecell[c]{\textbf{36}} & \makecell[c]{434} & \makecell[c]{99.48} & \makecell[c]{75.23}  \\
    \hline
    \makecell[c]{CWNNs~\cite{8641484}} & \makecell[c]{GRSL 2019} & \makecell[c]{81} & \makecell[c]{226} & \makecell[c]{99.66} & \makecell[c]{85.56}  \\
    \hline
    \makecell[c]{ESMOFCM~\cite{9134375}} & \makecell[c]{GRSL 2021} & \makecell[c]{95} & \makecell[c]{196} & \makecell[c]{\textbf{99.68}} & \makecell[c]{86.70}  \\
    \hline
    \makecell[c]{BIFLICM/D~\cite{DBLP:journals/lgrs/FangX22}} & \makecell[c]{GRSL 2022} & \makecell[c]{103} & \makecell[c]{718} & \makecell[c]{99.08} & \makecell[c]{\textbf{91.24}}  \\
    \hline
    \makecell[c]{ShearNet~\cite{9982693}} & \makecell[c]{TGRS 2022} & \makecell[c]{163} & \makecell[c]{\textbf{126}} & \makecell[c]{\textbf{99.68}} & \makecell[c]{87.41}  \\
    \hline\hline
  \end{tabular}}
  \label{tab:benchmark_SAR}
\end{table*}

%\subsubsection{Multispectral based Benchmarks}
\par \noindent \textbf{2) Multi-spectral Benchmarks.} Tab.~\ref{tab:benchmark_Multispectral} presents a collection of state-of-the-art models on several preeminent multi-spectral image datasets, including the LEVIR-CD dataset~\cite{Chen2020}, the CDD dataset~\cite{DBLP:journals/corr/abs-2108-07955}, the WHU Building dataset~\cite{8444434} and the SECOND dataset~\cite{DBLP:journals/tgrs/YangXLDYPZ22}. Specifically, in the LEVIR-CD dataset, P2V-CD~\cite{9975266} outperforms other methods and achieves the highest F1 score. ChangeStar~\cite{9709950} achieves the highest performance on the IoU metric, while ChangeFormer~\cite{9883686} performs the best on the OA metric. In the CDD dataset, P2V-CD~\cite{9975266} achieves the best performance of the F1 score, demonstrating superior results in both precision and recall metrics. Similarly, in the WHU Building dataset, P2V-CD~\cite{9975266} demonstrates remarkable performance, achieving the highest F1 score. DASNet~\cite{DBLP:journals/staeors/ChenYPCHZLL21} also exhibits competitive performance, obtaining the second-best F1 score. These findings indicate the effectiveness of P2V-CD and DASNet in the analysis of multi-spectral remote sensing data for building extraction tasks. In the SECOND dataset with multiple change detection categories, SSCD~\cite{DBLP:journals/tgrs/DingGLMZB22} achieves the highest score for the OA metric, while SCanNet~\cite{Ding2022JointSM} achieves the highest performance for the metrics mIoU, Sek, and $\text{F}_{scd}$.
\begin{table*}[!ht]
  \scriptsize
  \centering
  \caption{Some representative methods on Multi-spectral datasets. The model with the best performance is denoted in bold.}
  \scalebox{1.0}{
  \begin{tabular}
  {p{0.15\textwidth}|p{0.16\textwidth} |p{0.08\textwidth} |p{0.06\textwidth}|p{0.06\textwidth}
  |p{0.06\textwidth}|p{0.06\textwidth}}
    %\toprule %{c|l|m{7cm}}
    \hline\hline
  \rowcolor{gray!60!}  \multicolumn{7}{c} {\textbf{Multi-spctral Methods}}  \\
    \hline\hline
  \rowcolor{gray!30!}  \makecell[c]{\textbf{Method}} & \makecell[c]{\textbf{Source}} & \makecell[c]{\textbf{Precision}} & \makecell[c]{\textbf{Recall}} & \makecell[c]{\textbf{F1}} & \makecell[c]{\textbf{OA}} & \makecell[c]{\textbf{IoU}}\\  
    \hline\hline
  \rowcolor{gray!15!}   \multicolumn{7}{c}{\textbf{LEVIR\_CD Dataset}} \\
    \hline\hline  
    \makecell[c]{FC-EF~\cite{8451652}} & \makecell[c]{ICIP 2018} & \makecell[c]{90.64}  & \makecell[c]{78.84} & \makecell[c]{84.33} & \makecell[c]{98.39} & \makecell[c]{71.53}\\
    \hline
    \makecell[c]{STANet~\cite{Chen2020}} & \makecell[c]{Remote Sensing 2020} & \makecell[c]{83.81} & \makecell[c]{\textbf{91.00}} & \makecell[c]{87.26} & \makecell[c]{ 98.66} & \makecell[c]{77.40} \\
    \hline
    \makecell[c]{ChangeStar~\cite{9709950}} & \makecell[c]{ICCV 2021} & \makecell[c]{--} & \makecell[c]{--} & \makecell[c]{90.82} & \makecell[c]{--} & \makecell[c]{\textbf{83.19}} \\
    \hline
    \makecell[c]{ChangeFormer~\cite{9883686}} & \makecell[c]{IGARSS 2022} & \makecell[c]{92.05} & \makecell[c]{88.80} & \makecell[c]{90.40} & \makecell[c]{\textbf{99.04}} & \makecell[c]{82.48}  \\
    \hline
    \makecell[c]{BIT~\cite{9491802}} & \makecell[c]{TGRS 2022} & \makecell[c]{89.24} & \makecell[c]{89.37} & \makecell[c]{89.31} & \makecell[c]{98.92} & \makecell[c]{80.68} \\
    \hline
   \makecell[c]{SNUNet~\cite{9355573}} & \makecell[c]{GRSL 2022} & \makecell[c]{89.18} & \makecell[c]{87.17} & \makecell[c]{88.16} & \makecell[c]{98.82} & \makecell[c]{78.83} \\
    \hline
    \makecell[c]{P2V-CD~\cite{9975266}} & \makecell[c]{TIP 2023} & \makecell[c]{\textbf{93.32}} & \makecell[c]{90.60} & \makecell[c]{\textbf{91.94}} & \makecell[c]{--} & \makecell[c]{--} \\
    \hline\hline
    
  \rowcolor{gray!15!}   \multicolumn{7}{c}{\textbf{CDD Dataset}} \\
    \hline\hline
    \makecell[c]{FC-EF~\cite{8451652}} & \makecell[c]{ICIP 2018} & \makecell[c]{83.45} & \makecell[c]{\textbf{98.47}} & \makecell[c]{90.34} & \makecell[c]{97.58} & \makecell[c]{}  \\
    \hline
    \makecell[c]{STANet~\cite{Chen2020}} & \makecell[c]{Remote Sensing 2020} & \makecell[c]{95.17} & \makecell[c]{92.88} & \makecell[c]{94.01} & \makecell[c]{--} & \makecell[c]{} \\
    \hline
    \makecell[c]{ESCNet~\cite{9474911}} & \makecell[c]{TNNLS 2021} & \makecell[c]{90.04} & \makecell[c]{97.26} & \makecell[c]{93.51} & \makecell[c]{\textbf{98.45}} & \makecell[c]{} \\
    \hline
    \makecell[c]{ChangeFormer~\cite{9883686}} & \makecell[c]{IGARSS 2022} & \makecell[c]{94.50} & \makecell[c]{93.51} & \makecell[c]{94.23} & \makecell[c]{--} & \makecell[c]{} \\
    \hline
    \makecell[c]{BIT~\cite{9491802}} & \makecell[c]{TGRS 2022} & \makecell[c]{96.07} & \makecell[c]{93.49} & \makecell[c]{94.76} & \makecell[c]{--} & \makecell[c]{} \\
    \hline
    \makecell[c]{SNUNet~\cite{9355573}} & \makecell[c]{GRSL 2022} & \makecell[c]{98.09} & \makecell[c]{97.42} & \makecell[c]{97.75} & \makecell[c]{--} & \makecell[c]{} \\
    \hline
    \makecell[c]{DSAMNet~\cite{DBLP:journals/tgrs/ShiLLLWZ22}} & \makecell[c]{TGRS 2022} & \makecell[c]{94.54} & \makecell[c]{92.77} & \makecell[c]{93.69} & \makecell[c]{--} & \makecell[c]{} \\
    \hline
    \makecell[c]{P2V-CD~\cite{9975266}} & \makecell[c]{TIP 2023} & \makecell[c]{\textbf{98.57}} & \makecell[c]{98.26} & \makecell[c]{\textbf{98.42}} & \makecell[c]{--} & \makecell[c]{} \\
    \hline\hline
  
  \rowcolor{gray!15!}   \multicolumn{7}{c}{\textbf{WHU Building Dataset}} \\
    \hline\hline
    \makecell[c]{FC-EF~\cite{8451652}} & \makecell[c]{ICIP 2018} & \makecell[c]{71.63}  & \makecell[c]{67.25} & \makecell[c]{69.37} & \makecell[c]{97.61} & \makecell[c]{53.11}\\
    \hline
    \makecell[c]{STANet~\cite{Chen2020}} & \makecell[c]{Remote Sensing 2020} & \makecell[c]{79.37} & \makecell[c]{85.50} & \makecell[c]{82.32} & \makecell[c]{98.52} & \makecell[c]{69.95} \\
    \hline
    \makecell[c]{DASNet~\cite{DBLP:journals/staeors/ChenYPCHZLL21}} & \makecell[c]{J-STARS 2020} & \makecell[c]{90.00} & \makecell[c]{\textbf{90.50}} & \makecell[c]{91.00} & \makecell[c]{99.10} & \makecell[c]{--} \\
    \hline
    \makecell[c]{ChangeFormer~\cite{9883686}} & \makecell[c]{IGARSS 2022} & \makecell[c]{91.83} & \makecell[c]{88.02} & \makecell[c]{89.88} & \makecell[c]{\textbf{99.12}} & \makecell[c]{\textbf{81.63}}  \\
    \hline
    \makecell[c]{BIT~\cite{9491802}} & \makecell[c]{TGRS 2022} & \makecell[c]{86.64} & \makecell[c]{81.48} & \makecell[c]{83.98} & \makecell[c]{98.75} & \makecell[c]{72.39} \\
    \hline
   \makecell[c]{SNUNet~\cite{9355573}} & \makecell[c]{GRSL 2022} & \makecell[c]{89.90} & \makecell[c]{86.82} & \makecell[c]{88.33} & \makecell[c]{--} & \makecell[c]{--} \\
    \hline
    \makecell[c]{P2V-CD~\cite{9975266}} & \makecell[c]{TIP 2023} & \makecell[c]{\textbf{95.48}} & \makecell[c]{89.47} & \makecell[c]{\textbf{92.38}} & \makecell[c]{--} & \makecell[c]{--} \\
    \hline\hline

  \rowcolor{gray!15!}   \multicolumn{7}{c}{\textbf{SECOND Dataset}} \\
    \hline\hline  
  \rowcolor{gray!30!}  \makecell[c]{\textbf{Method}} & \makecell[c]{\textbf{Source}} & \makecell[c]{\textbf{OA}} & \makecell[c]{\textbf{mIoU}} & \makecell[c]{\textbf{Sek}} & \makecell[c]{$\boldsymbol{F_{scd}}$}  \\  
    \hline\hline
    \makecell[c]{HRSCD~\cite{Daudt2018MultitaskLF}} & \makecell[c]{CVIU 2019} & \makecell[c]{86.62}  & \makecell[c]{71.15} & \makecell[c]{18.80} & \makecell[c]{58.21} 
  \\
    \hline
    \makecell[c]{ASN~\cite{DBLP:journals/tgrs/YangXLDYPZ22}} & \makecell[c]{TGRS 2021} & \makecell[c]{--} & \makecell[c]{69.50} & \makecell[c]{16.30} & \makecell[c]{--} \\
    \hline
    \makecell[c]{SSCD~\cite{DBLP:journals/tgrs/DingGLMZB22}} & \makecell[c]{TGRS 2022} & \makecell[c]{87.19} & \makecell[c]{72.60} & \makecell[c]{21.86} & \makecell[c]{61.22}   \\
    \hline
    \makecell[c]{SSESN~\cite{DBLP:journals/staeors/ZhaoZGLYXL22}} & \makecell[c]{J-STARS 2022} & \makecell[c]{\textbf{89.00}} & \makecell[c]{70.80} & \makecell[c]{--} & \makecell[c]{--}  \\
    \hline
    \makecell[c]{SCanNet~\cite{Ding2022JointSM}} & \makecell[c]{ArXiv 2022} & \makecell[c]{87.76} & \makecell[c]{\textbf{73.42}} & \makecell[c]{\textbf{23.94}} & \makecell[c]{\textbf{63.66}}  \\
    \hline\hline
   \end{tabular}}
  \label{tab:benchmark_Multispectral}
\end{table*}  

  %\subsubsection{Hyperspectral based Benchmarks}
  \par \noindent \textbf{3) Hyperspectral Benchmarks.} In Tab.~\ref{tab:benchmark_Hyperspectral}, we present some representative methods on two dominant hyperspectral datasets, i.e. the River dataset~\cite{Wang2019GETNETAG} and the Hermiston dataset~\cite{DBLP:journals/tgrs/WangWWWB22}. Specifically, in the River dataset, MSDFFN~\cite{DBLP:journals/tgrs/LuoZLGGR23} outperforms other methods with respect to the Recall, F1 score, OA, and KC metrics. Meanwhile, SSA-SiamNet~\cite{9494085} achieves the highest precision score compared to all other evaluated methods. In the Hermiston dataset, MSDFFN~\cite{DBLP:journals/tgrs/LuoZLGGR23} achieves the best performance in terms of precision, recall, F1 score, and OA metrics. Conversely, GETNET~\cite{Wang2019GETNETAG} achieves the best results in the KC metric compared to other methods.
\begin{table*}[!ht]
  \scriptsize
  \centering
  \caption{Some representative methods on hyperspectral datasets. The model with the best performance is denoted in bold.}
  \scalebox{1.0}{
  \begin{tabular}
  {p{0.15\textwidth}|p{0.10\textwidth} |p{0.08\textwidth} |p{0.06\textwidth}|p{0.06\textwidth}
  |p{0.06\textwidth}|p{0.06\textwidth}}
    %\toprule %{c|l|m{7cm}}
    \hline\hline

  \rowcolor{gray!60!}  \multicolumn{7}{c} {\textbf{Hyperspctral Methods}}  \\
    \hline\hline
  \rowcolor{gray!30!}  \makecell[c]{\textbf{Method}} & \makecell[c]{\textbf{Source}} & \makecell[c]{\textbf{Precision}} & \makecell[c]{\textbf{Recall}} & \makecell[c]{\textbf{F1}} & \makecell[c]{\textbf{OA}} & \makecell[c]{\textbf{KC}}\\  
    \hline\hline
  \rowcolor{gray!15!}   \multicolumn{7}{c}{\textbf{River Dataset}} \\
    \hline\hline  
   % \makecell[c]{CVA~\cite{1980Change}} & \makecell[c]{1980} & \makecell[c]{} & \makecell[c]{61.56}  & \makecell[c]{89.16} & \makecell[c]{72.84} & \makecell[c]{94.22} & \makecell[c]{69.72}\\
    %\hline
    %\makecell[c]{SVM~\cite{Nemmour2006MultipleSV}} & \makecell[c]{ISPRS 2006} & \makecell[c]{90.17} & \makecell[c]{69.79} & \makecell[c]{78.68} & \makecell[c]{96.71} & \makecell[c]{76.93} \\
    %\hline
    \makecell[c]{PCA-CVA~\cite{DBLP:journals/lgrs/ZhangGZSS16}} & \makecell[c]{GRSL 2016} & \makecell[c]{--} & \makecell[c]{--} & \makecell[c]{--} & \makecell[c]{95.16} & \makecell[c]{74.77} \\
    \hline
    \makecell[c]{GETNET~\cite{Wang2019GETNETAG}} & \makecell[c]{TGRS 2018} & \makecell[c]{85.64} & \makecell[c]{78.98} & \makecell[c]{82.18} & \makecell[c]{97.18} & \makecell[c]{80.53}  \\
    \hline
    \makecell[c]{SiamCRNN~\cite{DBLP:journals/tgrs/ChenWDZW20}} & \makecell[c]{TGRS 2020} & \makecell[c]{88.14} & \makecell[c]{69.12} & \makecell[c]{77.45} & \makecell[c]{96.5} & \makecell[c]{75.59} \\
    \hline
    \makecell[c]{SSA-SiamNet~\cite{9494085}} & \makecell[c]{TGRS 2021} & \makecell[c]{\textbf{91.89}} & \makecell[c]{74.10} & \makecell[c]{82.04} & \makecell[c]{97.18} & \makecell[c]{80.53} \\
    \hline
    \makecell[c]{ML-EDAN~\cite{DBLP:journals/tgrs/QuHDLX22}} & \makecell[c]{TGRS 2022} & \makecell[c]{89.57} & \makecell[c]{83.75} & \makecell[c]{86.57} & \makecell[c]{97.74} & \makecell[c]{85.33} \\
    \hline
   \makecell[c]{MSDFFN~\cite{DBLP:journals/tgrs/LuoZLGGR23}} & \makecell[c]{TGRS 2023} & \makecell[c]{90.52} & \makecell[c]{\textbf{87.58}} & \makecell[c]{\textbf{89.01}} & \makecell[c]{\textbf{98.12}} & \makecell[c]{\textbf{87.98}} \\
    \hline\hline
    
  \rowcolor{gray!15!}   \multicolumn{7}{c}{\textbf{Hermiston Dataset}} \\
    \hline\hline
%\makecell[c]{CVA~\cite{1980Change}} & \makecell[c]{1980} & \makecell[c]{} & \makecell[c]{\textbf{97.61}}  & \makecell[c]{74.63} & \makecell[c]{84.59} & \makecell[c]{93.87} & \makecell[c]{80.85}\\
 %   \hline
  %  \makecell[c]{SVM~\cite{Nemmour2006MultipleSV}} & \makecell[c]{ISPRS 2006} & \makecell[c]{94.17} & \makecell[c]{88.03} & \makecell[c]{91.00} & \makecell[c]{96.08} & \makecell[c]{88.49} \\
  %  \hline
    \makecell[c]{GETNET~\cite{Wang2019GETNETAG}} & \makecell[c]{TGRS 2018} & \makecell[c]{92.99} & \makecell[c]{90.16} & \makecell[c]{91.50} & \makecell[c]{89.09} & \makecell[c]{\textbf{96.23}}  \\
    \hline
    \makecell[c]{SiamCRNN~\cite{DBLP:journals/tgrs/ChenWDZW20}} & \makecell[c]{TGRS 2020} & \makecell[c]{92.66} & \makecell[c]{49.28} & \makecell[c]{62.67} & \makecell[c]{87.35} & \makecell[c]{56.15} \\
    \hline
    \makecell[c]{SSA-SiamNet~\cite{9494085}} & \makecell[c]{TGRS 2021} & \makecell[c]{93.18} & \makecell[c]{89.17} & \makecell[c]{91.45} & \makecell[c]{96.22} & \makecell[c]{89.02} \\
    \hline
    \makecell[c]{RSCNet~\cite{DBLP:journals/tgrs/WangWWB22}} & \makecell[c]{TGRS 2022} & \makecell[c]{93.98} & \makecell[c]{91.32} & \makecell[c]{92.63} & \makecell[c]{96.73} & \makecell[c]{90.53} \\
    \hline
    \makecell[c]{ML-EDAN~\cite{DBLP:journals/tgrs/QuHDLX22}} & \makecell[c]{TGRS 2022} & \makecell[c]{94.88} & \makecell[c]{92.53} & \makecell[c]{93.68} & \makecell[c]{97.19} & \makecell[c]{91.87} \\
    \hline
   \makecell[c]{MSDFFN~\cite{DBLP:journals/tgrs/LuoZLGGR23}} & \makecell[c]{TGRS 2023} & \makecell[c]{\textbf{95.55}} & \makecell[c]{\textbf{93.69}} & \makecell[c]{\textbf{94.61}} & \makecell[c]{\textbf{97.59}} & \makecell[c]{93.06} \\
    \hline\hline
  \end{tabular}}
  \label{tab:benchmark_Hyperspectral}
\end{table*}   

  %\subsubsection{Heterogeneous based Benchmarks}
  \par \noindent \textbf{4) Heterogeneous Benchmarks.} Tab.~\ref{tab:benchmark_Heterogenous} presents several dominant methods that have been applied to the heterogeneous dataset, i.e. the California dataset~\cite{DBLP:journals/tnn/YangJLHYJ22}. Specifically, DPFL-Net-4~\cite{DBLP:journals/tnn/YangJLHYJ22} demonstrates superior performance compared to other methods when evaluated based on the OA, KC, and AUC metrics. Notably, it achieves state-of-the-art results in all three metrics and particularly outperforms other methods by a significant margin in the KC metric.
\begin{table}[!ht]
  \scriptsize
  \centering
  \caption{Some representative methods on heterogeneous datasets. The model with the best performance is denoted in bold.}
  \scalebox{1.0}{
  \begin{tabular}
  {p{0.1\textwidth}|p{0.08\textwidth} |p{0.04\textwidth} |p{0.04\textwidth}|p{0.04\textwidth}
  |p{0.04\textwidth}}
    %\toprule %{c|l|m{7cm}}
    \hline\hline

  \rowcolor{gray!60!}  \multicolumn{6}{c} {\textbf{Heterogeneous Methods}}  \\
    \hline\hline
  \rowcolor{gray!30!}  \makecell[c]{\textbf{Method}} & \makecell[c]{\textbf{Source}} & \makecell[c]{\textbf{OA}} & \makecell[c]{\textbf{KC}} & \makecell[c]{\textbf{AUC}} & \makecell[c]{\textbf{F1}}  \\  
    \hline\hline
  \rowcolor{gray!15!}   \multicolumn{6}{c}{\textbf{California Dataset}} \\
    \hline\hline  
    \makecell[c]{SCCN~\cite{DBLP:journals/tnn/LiuGQZ18}} & \makecell[c]{TNNLS 2018} & \makecell[c]{97.60}  & \makecell[c]{87.17} & \makecell[c]{98.78} & \makecell[c]{--} 
 \\
    \hline
    \makecell[c]{AM\_HPT~\cite{8798991}} & \makecell[c]{TGRS 2019} & \makecell[c]{98.12} & \makecell[c]{90.18} & \makecell[c]{99.24} & \makecell[c]{--}  \\
    \hline
    \makecell[c]{CAN~\cite{DBLP:journals/lgrs/NiuGZY19}} & \makecell[c]{GRSL 2019} & \makecell[c]{90.40} & \makecell[c]{36.50} & \makecell[c]{--} & \makecell[c]{42.40}  \\
    \hline
    \makecell[c]{ACE-Net~\cite{DBLP:journals/tgrs/LuppinoKBMSJA22}} & \makecell[c]{TGRS 2021} & \makecell[c]{91.50} & \makecell[c]{41.50} & \makecell[c]{--} & \makecell[c]{\textbf{45.90}}  \\
    \hline
    \makecell[c]{CA\_AE~\cite{9773305}} & \makecell[c]{TNNLS 2022} & \makecell[c]{97.88} & \makecell[c]{88.66} & \makecell[c]{99.18} & \makecell[c]{--}  \\
    \hline
   \makecell[c]{DPFL-Net-4~\cite{DBLP:journals/tnn/YangJLHYJ22}} & \makecell[c]{TNNLS 2022} & \makecell[c]{\textbf{98.89}} & \makecell[c]{\textbf{94.17}} & \makecell[c]{\textbf{99.79}} & \makecell[c]{--}  \\
    \hline\hline
  \end{tabular}}
  \label{tab:benchmark_Heterogenous}
\end{table}

%\subsubsection{Point cloud based Benchmarks}
\par \noindent \textbf{5) 3D Change Detection Benchmarks.} Tab.~\ref{tab:benchmark_3D} presents some representative models on 3D change detection dataset, i.e. the 3DCD dataset~\cite{MARSOCCI2023325}. To be specific, we find that MTBIT~\cite{MARSOCCI2023325} achieves state-of-the-art performance on both RMSE and cRMSE metrics.
\begin{table}[!ht]
  \scriptsize
  \centering
  \caption{ Some representative methods on 3D change detection datasets. The model with the best performance is denoted in bold.}
  \scalebox{1.0}{
  \begin{tabular}
  {p{0.15\textwidth}|p{0.10\textwidth} |p{0.06\textwidth} |p{0.06\textwidth}}
    %\toprule %{c|l|m{7cm}}
    \hline\hline

  \rowcolor{gray!60!}  \multicolumn{4}{c} {\textbf{3D Change Detection Methods}}  \\
    \hline\hline
  \rowcolor{gray!30!}  \makecell[c]{\textbf{Method}} & \makecell[c]{\textbf{Source}} & \makecell[c]{\textbf{RMSE}} & \makecell[c]{\textbf{cRMSE}}  \\  
    \hline\hline
  \rowcolor{gray!15!}   \multicolumn{4}{c}{\textbf{3DCD Dataset}} \\
    \hline\hline  
    \makecell[c]{IM2HEIGHT~\cite{Mou2018IM2HEIGHTHE}} & \makecell[c]{ArXiv 2018} & \makecell[c]{1.57}  & \makecell[c]{7.59} 
 \\
    \hline
    \makecell[c]{FC-EF~\cite{8451652}} & \makecell[c]{ICIP 2018} & \makecell[c]{1.41} & \makecell[c]{7.04}  \\
    \hline
    \makecell[c]{ChangeFormer~\cite{9883686}} & \makecell[c]{IGARSS 2022} & \makecell[c]{1.31} & \makecell[c]{7.09}   \\
    \hline
    \makecell[c]{SNUNet~\cite{9355573}} & \makecell[c]{GRSL 2022} & \makecell[c]{1.24} & \makecell[c]{6.47}  \\
    \hline
    \makecell[c]{MTBIT~\cite{MARSOCCI2023325}} & \makecell[c]{ISPRS 2023} & \makecell[c]{\textbf{1.20}} & \makecell[c]{\textbf{6.46}}  \\
    \hline\hline
  \end{tabular}}
  \label{tab:benchmark_3D}
\end{table}
\section{Future Trends} % one page 
\label{future}
\par \noindent \textbf{Generalization of the change detection algorithms:} The generalization performance of change detection algorithms in remote sensing is critical for their practical applications, as they need to be capable of effectively detecting changes in new and unobserved data, diverse geographical locations, different sensors, and varying environmental conditions. To improve their generalization, future trends include the use of transfer learning to transfer knowledge learned from one task or domain to others, domain adaptation~\cite{zhou2022context} aims to adapt models to different environmental conditions or sensors, multi-sensor fusion to integrate data from different sensors, and explainable AI to interpret the decision-making process of a model. These trends are expected to enhance the generalization and overall performance of change detection algorithms in remote sensing images.

\par \noindent \textbf{Learning with few samples:} Few-shot learning~\cite{catalano2023few,han2023reference} aims to train models with only a small amount of examples for each category, enabling them to generalize to new classes. The future trends in few-shot learning for change detection algorithms in remote sensing images include several key approaches, such as meta-learning, generative models, and domain generalization. Meta-learning involves enabling models to learn how to learn and quickly adapt to new classes. This approach allows models to leverage their past learning experiences and generalize to new tasks with fewer examples. Domain generalization~\cite{zhou2022domain} enables models to perform well on new and unseen domains. This approach involves training the model on data from different domains to improve its ability to generalize to new environments. By adapting to new geographical locations, sensors, or environmental conditions, these trends will enable algorithms to detect new types of changes with few examples, leading to better decision-making and a deeper understanding of the environment.

\par \noindent \textbf{Deep dive in the transformer-based algorithms:} Transformer-based models~\cite{carion2020end,xu2022fashionformer,li2023panopticpartformer++,li2023tube,RT_DETR,li2022video,Transformer_survey} have achieved significant progress in computer vision. In particular, the self-attention-based methods~\cite{Li_2021_CVPR,DBLP:conf/iclr/DosovitskiyB0WZ21,wang2023convolution} achieve better results than pure convolution-based methods for representation learning.
The development of transformer-based algorithms for change detection in remote sensing images represents an emerging area of research with significant potential. To shape the future development of these algorithms, several trends have been identified, including the enhancement of attention mechanisms to better capture complex patterns and dependencies, integration with other deep learning models to create more robust and accurate change detection systems, and the use of unsupervised and semi-supervised learning approaches to develop robust and accurate transformer models that can learn from unlabeled or partially labeled data. Additionally, future research will focus on the development of transformer models that can handle and integrate different data modalities, such as optical, radar, and LiDAR, to enhance the accuracy of change detection. Finally, online and continual learning approaches~\cite{DBLP:conf/cvpr/DouillardCDC21, DBLP:journals/tgrs/FengSDLGF22} will enable transformer models to learn and adapt to new data streams, allowing for more accurate and robust change detection systems.

\par \noindent \textbf{Efficient models for practical applications:} Efficiency is a crucial factor in developing practical and scalable change detection algorithms for remote sensing images. Here are some of the future trends in the development of efficient models for change detection: \textit{Sparse and lightweight models}~\cite{ liu2021paddleseg,li2020semantic,zhang2023rethinking,SRNetxt,wan2023seaformer}. To reduce the computational complexity and memory footprint of change detection algorithms, future research will focus on developing sparse and lightweight models. These models will be designed to perform the change detection task with a minimal number of parameters and operations while maintaining high accuracy. \textit{Compression techniques}~\cite{DBLP:journals/tgrs/ChenWCCQ22, DBLP:journals/tgrs/ZhangYSDFW22}: Compression techniques such as pruning, quantization, and knowledge distillation can reduce the size of deep learning models without sacrificing performance. Future research will explore ways to apply these techniques to change detection models, reducing their memory and computational requirements.

\par \noindent \textbf{Generate synthetic data for joint training:} To train a high-performing and generalized change detection model, a significant amount of labeled data is required, which in turn demands substantial manual efforts for data collection and annotation. One possible solution is to employ a generated synthetic dataset. Recently, diffusion-based generation models~\cite{DBLP:conf/nips/DhariwalN21, DBLP:journals/corr/abs-2303-11681} have provided the opportunity to create higher-quality images and corresponding masks than previous generative models. Generative images with diffusion models exhibit fewer domain gaps than real images, offering greater potential for training change detection models without reliance on real data. Furthermore, depending on the specific application scenarios, generative images, and masks can be produced to cater to task-specific needs, including few-shot or long-tail tasks.

\par \noindent \textbf{Integration of multi-source/modal and multiple datasets:} As delineated in the preliminary section, numerous datasets exist for each data source. A plausible approach to enhancing model performance and generality is amalgamating these datasets for each data source to create a more extensive change detection dataset. It is important to recognize that dissimilarities may exist among datasets from different domains. Thus, minimizing these domain gaps represents another research concern. Additionally, further research could be conducted to investigate cross-domain multi-source datasets~\cite{cai2022bigdetection} in order to improve model training. For instance, one could explore methods for distilling the knowledge~\cite{zhao2023learn} acquired from a vast amount of multi-spectral change data into secrecy SAR data, thereby achieving superior performance on corresponding SAR data. One potential avenue for future research is to explore the integration of multi-modal data, specifically the incorporation of image and text pairs, to improve the accuracy of selected area change detection. It is possible that the creation of a large-scale image-text based change detection dataset could facilitate progress in this direction. This approach could provide a more comprehensive understanding of changes occurring in a given area by incorporating both visual and textual data. The utilization of multi-modal data has been shown to be effective in other areas of research, and it is possible that this approach could yield promising results in the field of selected area change detection. Further investigation in this direction may help to identify new patterns and insights related to selected area change detection.

\par \noindent \textbf{Exploration based on foundation models:} Recently, there has been a surge in the development of foundation models, such as Segment Anything~\cite{kirillov2023segany}, Painter~\cite{Painter}, and SegGPT~\cite{SegGPT}, that leverage large amounts of data and GPU resources. For example, Segment Anything, also known as SAM, is a highly effective method for segmenting all types of targets, making it highly versatile for remote sensing applications, especially for high-resolution images. To further improve change detection tasks, there are two promising research avenues. First, an automatic approach can be developed to generate a more extensive change detection dataset, which can help to train more comprehensive aerial-specific foundation models. Second, integrating foundation models can address domain adaptation gaps and significantly enhance change detection performance. 

\section{Conclusions}
\label{conclusion}
This paper has provided a comprehensive and in-depth survey of the recent advancements in change detection for remote sensing images, which have been achieved over the past decade. Through the comprehensive review of fundamental knowledge and the classification of existing algorithms, this paper has provided a detailed and organized understanding of the current state of the field. Additionally, the summary of the state-of-the-art performance on several datasets has demonstrated the effectiveness of deep learning techniques in addressing the challenges of change detection. Finally, the identification of future research directions has provided valuable insights into the potential avenues for further advancement of the field. It is our sincere hope that this survey paper will not only contribute to the current understanding of change detection in remote sensing but also inspire and guide future research efforts in this area.

\bibliographystyle{IEEEtran}
\bibliography{main} % at least 300 references 5 pages 
%\onecolumn

\end{document}